\documentclass{article}
\usepackage{iclr2027_conference,times}

\iclrfinalcopy

\usepackage[utf8]{inputenc}
\usepackage[T1]{fontenc}
\usepackage[table]{xcolor}
\usepackage{amsmath,amssymb,amsfonts,amsthm}
\usepackage{algorithm}
\usepackage{algpseudocode}
\usepackage{booktabs}
\usepackage{enumitem}
\usepackage{float}
\usepackage{graphicx}
\usepackage{listings}
\usepackage{mdframed}
\usepackage{microtype}
\usepackage{multirow}
\usepackage{nicefrac}
\usepackage[most,skins,theorems]{tcolorbox}
\usepackage{tikz}
\usepackage{wrapfig}
\usepackage{url}
\usepackage[hidelinks]{hyperref}

\usetikzlibrary{positioning,arrows.meta,calc,shapes.geometric,fit,backgrounds}
\algtext*{EndFor}
\algtext*{EndIf}

\lstdefinestyle{promptstyle}{
  basicstyle=\ttfamily\footnotesize\linespread{0.95}\selectfont,
  breaklines=true,
  columns=fullflexible,
  keepspaces=true,
  showstringspaces=false,
  escapeinside={||},
  frame=none,
  xleftmargin=0pt,
  xrightmargin=0pt,
}

\definecolor{lightgray}{gray}{0.93}
\definecolor{darkred}{RGB}{160,0,0}
\definecolor{darkrowgreen}{RGB}{205,232,215}
\definecolor{mygreen}{RGB}{35,120,45}
\definecolor{myred}{RGB}{180,30,40}
\definecolor{lightgraytext}{RGB}{170,170,170}
\definecolor{opsdblue}{RGB}{75,125,160}
\definecolor{grpogreen}{RGB}{75,130,75}
\definecolor{lightorange}{HTML}{faa755}
\definecolor{lightblue}{RGB}{220,235,250}
\definecolor{bsdcol}{RGB}{31,85,167}

\newcommand{\ours}{CriPO}
\newcommand{\oursfull}{Criterion-Distilled Policy Optimization}
\newcommand{\uprise}[1]{\textcolor{mygreen}{\small\;{$\uparrow$ #1}}}
\newcommand{\down}[1]{\textcolor{myred}{\small\;{$\downarrow$ #1}}}
\newcommand{\basex}[1]{\textcolor{lightgraytext}{\small\;{$\uparrow$ #1}}}
\newcommand{\best}[1]{\textbf{#1}}
\newcommand{\second}[1]{\underline{#1}}

\tcbset{
  takeawaysbox/.style={
    title=Takeaways,
    colback=lightblue!80,
    colframe=black,
    fonttitle=\bfseries\small,
    coltitle=white,
    colbacktitle=black,
    enhanced,
    attach boxed title to top left={xshift=2.5mm,yshift=-2.5mm},
    boxed title style={rounded corners,size=small,colframe=black,colback=black},
    width=\linewidth,
    arc=3.5mm
  }
}

\title{\ours{}: Enhancing Rubric-based RL via Self-Distillation}
\author{
\textbf{Mingxuan Xia\textsuperscript{1,2}}\thanks{Equal contribution. Work done during internship at ByteDance.}\quad
\textbf{Yuhang Yang\textsuperscript{1,2}}\footnotemark[1]\quad
\textbf{Chao Ye\textsuperscript{2}}\quad
\textbf{Shuai Zhu\textsuperscript{2}}\quad
\textbf{Shenzhi Yang\textsuperscript{1}}
\\[0.3em]
\textbf{Guangcheng Zhu\textsuperscript{1}}\quad
\textbf{Yuhang Zhang\textsuperscript{2}}\quad
\textbf{Cheng Peng\textsuperscript{1}}\quad
\textbf{Haobo Wang\textsuperscript{1}}\thanks{Corresponding author.}\quad
\textbf{Chenglong Wang\textsuperscript{2}}\footnotemark[2]
\\[0.4em]
\textsuperscript{1}Zhejiang University\quad
\textsuperscript{2}ByteDance \\[0.4em]
\small{\texttt{\{xiamingxuan,yangyuhang,wanghaobo\}@zju.edu.cn}} \\
\small{\texttt{\{yechao.66,wangsiqing.jacky\}@bytedance.com}}
}

\hypersetup{
  pdftitle={\ours{}: Enhancing Rubric-based RL via Self-Distillation},
  pdfauthor={Mingxuan Xia, Yuhang Yang, Chao Ye, Shuai Zhu, Shenzhi Yang, Guangcheng Zhu, Yuhang Zhang, Cheng Peng, Haobo Wang, Chenglong Wang}
}

\begin{document}
\maketitle
\lhead{Preprint}

\begin{abstract}
Rubric-based Reinforcement Learning (RL) has recently shown promise in improving Large Language Models (LLMs) on open-ended tasks. A widely recognized limitation of rubric-based RL is limited exploration: criteria that no rollout manages to satisfy (\textit{Unexplored Criteria}) receive no optimization signal. Recent methods address this by incorporating rubric information as external guidance during rollout generation, yet they introduce a \textit{train-inference mismatch}: the policy is optimized on rollouts produced under external guidance while this guidance is absent at inference time, causing error accumulation through autoregressive decoding. Moreover, these exploration-focused approaches overlook a fundamentally different failure mode that we term \textit{Suppressed Criteria}---criteria that are satisfied by some rollouts yet whose learning signals might be lost during optimization because scalar reward aggregation assigns them non-positive aggregate advantages. Our analysis reveals that suppressed criteria constitute a persistent and non-negligible failure mode---over 25\% of samples exhibit this issue throughout training. To simultaneously address both unexplored and suppressed criteria without introducing training-inference mismatch, we propose \textbf{\oursfull{} (\ours{})}, which enhances rubric-based RL via on-policy self-distillation. For unexplored criteria, \ours{} constructs a behavior-injection teacher and computes a filtered forward-KL loss to inject missing behaviors into the policy. For suppressed criteria, \ours{} uses a counterfactual teacher to locate criterion-relevant tokens in negative-advantage rollouts, and corrects their advantages in GRPO to preserve useful patterns.
Experiments on medicine and science benchmarks demonstrate that \ours{} outperforms existing rubric-based RL methods, e.g., achieving an average gain of \textbf{3.3} points over GRPO on Qwen3-4B.
\end{abstract}

\section{Introduction}
\label{sec:introduction}

In recent years, Reinforcement Learning from Verifiable Rewards (RLVR), instantiated by methods such as GRPO~\citep{grpo}, has emerged as a prominent paradigm for improving the reasoning capabilities of large language models~\citep{deepseek-r1,prime}. RLVR has achieved notable success in verifiable domains such as mathematical reasoning and code generation, where exact answers or executable test cases provide stable and reliable supervision~\citep{OlympiadBench, SWE-RL}. However, in open-ended scenarios such as medical consultation and long-form writing, model outputs typically do not admit a unique gold answer and are difficult to verify with deterministic rules. To extend RLVR to such domains, recent work has explored rubric-based rewards, which decompose response quality into interpretable evaluation criteria, use LLM-as-a-Judge~\citep{llm_as_judge} to assign multi-dimensional scores, and aggregate them into a scalar reward for RL training~\citep{rubrics_as_rewards, checklists, qalign, rules}.

While rubric-based RL has shown promising results in these open-ended domains, it still faces a fundamental challenge of limited exploration: since GRPO only optimizes over behaviors present in the sampled rollout group~\citep{bounded_exploration,Imbalanced_optimization}, criteria that no rollout manages to satisfy, termed \textbf{\textit{Unexplored Criteria}}, receive no effective optimization signal. 
Recent methods address this issue by using rubrics as external guidance during rollout generation. For example, RuscaRL~\citep{ruscarl} uses rubric criteria as explicit scaffolding to elicit more diverse and higher-quality rollouts, while HeRL~\citep{herl} leverages failed trajectories and their unmet criteria as hindsight guidance for generating revised rollouts.
However, these exploration-enhancement methods still have two important drawbacks. First, they introduce a \textbf{\textit{training-inference mismatch}}: during training, rollouts are generated with privileged information, so each token is optimized under externally guided prefixes; at inference time, however, the model must generate solely from its own prefixes. This discrepancy constitutes exposure bias~\citep{opd,opd_survey,rethinkopd}, where early generation error that could be avoided during guided training may propagate through subsequent autoregressive decoding.
Also, these mixed-policy optimization pipelines further lead to longer training times; see Section \ref{app:compute_cost} for more details.

\begin{figure}[t]
\centering
\includegraphics[width=\textwidth]{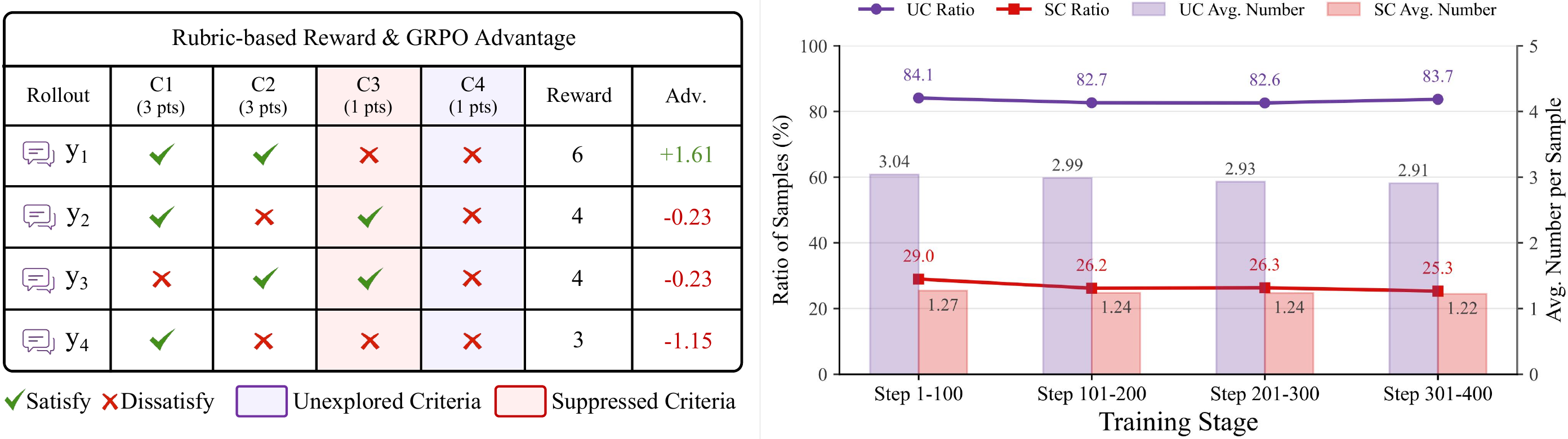}
\caption{\textbf{Left:} Two failure modes in GRPO: \textit{Unexplored Criteria} (UC), which are never satisfied by any rollout in the group (C4), and \textit{Suppressed Criteria} (SC), for which criterion-satisfying rollouts receive non-positive aggregate advantages (C3). \textbf{Right:} Ratio of samples with UC/SC and number of UC/SC on average among samples containing UC/SC when training Qwen3-4B on RaR-Medicine.}
\label{fig:blind_criteria}
\end{figure}

Second, these methods overlook another fundamentally different failure mode beyond unexplored criteria: \textbf{\textit{Suppressed Criteria}}---already explored criteria whose learning signals might be systematically lost during GRPO optimization. This occurs because GRPO aggregates criterion-wise scores into a single scalar reward and broadcasts it uniformly to the entire response.
Consequently, rollouts that satisfy certain criteria but underperform on others receive negative or negligible aggregate advantages, causing useful criterion-satisfying behaviors to be penalized or ignored rather than reinforced. For example, in Figure~\ref{fig:blind_criteria} (left), $y_2$ and $y_3$ satisfy criterion C3, but their advantage sum is negative (-0.46).
Our statistical analysis in Figure~\ref{fig:blind_criteria} (right) reveals that suppressed criteria remain a non-negligible issue in practice: when training Qwen3-4B on RaR-Medicine, over 25\% of samples contain suppressed criteria throughout training, with more than 1.2 suppressed criteria on average among samples containing SC.
While some prior works~\citep{rucl,alternating,rtt} also recognize that scalar reward aggregation can obscure individual criterion contributions, they do not provide a systematic analysis of when and why criterion-level signals are lost, and they also fail to address the complementary challenge of unexplored criteria.

To address both unexplored and suppressed criteria without introducing training-inference mismatch, in this paper, we propose \textbf{\oursfull{} (\ours{})}, which enhances rubric-based RL via On-Policy Self-Distillation (OPSD)~\citep{opsd,sdpo,opcd}. 
Unlike the approaches that inject knowledge or behaviors by producing rollouts under privileged information, OPSD constructs a privilege-conditioned self-teacher and provides token-level supervision on the model's own on-policy rollouts, thereby obtaining learning signals without a distributional gap between training and inference.
Specifically, \ours{} adopts GRPO as the stable, reward-grounded optimization backbone and uses the following two OPSD-derived signals for targeted criterion-level correction: for unexplored criteria, \ours{} constructs a \textit{behavior-injection self-teacher} and computes forward-KL loss with contribution-based token filtering as an auxiliary loss to inject missing criterion-specific behaviors into the policy; for suppressed criteria, \ours{} constructs a \textit{counterfactual self-teacher} to locate suppressed-criterion-relevant tokens in negative-advantage rollouts, and flip or mask their GRPO advantages to preserve useful patterns.
Experiments on five medicine and science benchmarks demonstrate that \ours{} outperforms five baselines in both in-domain and cross-domain settings (e.g., achieving an average gain of \textbf{3.3} points over GRPO on Qwen3-4B), and further analysis shows that \ours{} effectively reduces the occurrence rates of both unexplored and suppressed criteria.
Note that \ours{} requires at most one additional forward pass per rollout, resulting in about a 7\% increase on average in training time compared with GRPO, and \ours{} also improves optimization efficiency, reaching GRPO's converged performance in no more than half the optimization steps and less than half the wall-clock training time.

\section{Preliminaries}\label{sec:preliminary}

\paragraph{Rubric-based Reinforcement Learning (RL).} We adopt Group Relative Policy Optimization (GRPO)~\citep{grpo} as the core RL algorithm for training LLMs with rubric-based rewards.
Given a prompt $x$, the policy $\pi_\theta$ samples a group of on-policy rollouts $\{y_i\}_{i=1}^{G}$, where $y_i\sim\pi_\theta(\cdot\mid x)$. Each rollout is evaluated with a set of criteria $\mathcal{C}=\{(c_j,\omega_j)\}_{j=1}^{M}$, where $c_j$ denotes an evaluation criterion and $\omega_j$ its importance weight. A judge model assigns a criterion-wise score $r_{ij}=c_j(x,y_i)$, indicating how well rollout $y_i$ satisfies criterion $c_j$. These criterion-wise scores are then aggregated into a scalar reward: $R_i=\sum_{j=1}^{M}\omega_j r_{ij} / \sum_{j=1}^{M}\omega_j$.
GRPO then computes a group-relative advantage by normalizing the reward within the group: $A_i=(R_i-\mu_R)/(\sigma_R+\epsilon)$, where $\mu_R$ and $\sigma_R$ are the mean and standard deviation of the group's reward, and $\epsilon>0$ avoids division by zero. The policy model is optimized with the PPO-style~\citep{ppo} clipped surrogate objective:
\begin{equation}
    \mathcal{L}_{\mathrm{GRPO}}
    =
    -\mathbb{E}_{x,\{y_i\}_{i=1}^G\sim\pi_{\theta_\mathrm{old}}(\cdot|x)}\left[\frac{1}{\sum_i|y_i|}\sum_{i=1}^{G}
    \sum_{t=1}^{|y_i|}
    \min\!\left(
    \rho_{i,t}A_i,
    \mathrm{clip}\left(\rho_{i,t},1-\epsilon_{\mathrm{c}},1+\epsilon_{\mathrm{c}}\right)A_i
    \right)\right],
\end{equation}
where $\rho_{i,t}=\pi_\theta(y_{i,t}\mid x,y_{i,<t})/
\pi_{\theta_{\mathrm{old}}}(y_{i,t}\mid x,y_{i,<t})$ is the importance ratio and $\pi_{\theta_\mathrm{old}}$ is the frozen rollout policy and $\pi_\theta$ is the current policy with gradient. Note that we use token-mean aggregation following the default implementation of GRPO in the verl codebase \citep{verl}.


\paragraph{Failure in Rubric-based RL: Unexplored and Suppressed Criteria.}
Although rubric-based rewards provide criterion-level supervision, two inherent limitations of GRPO prevent this supervision from being fully exploited: (i) \textit{limited exploration}: GRPO only optimizes over behaviors present in the sampled rollouts, so that criteria with no rollout managed to satisfy receive no optimization signal; and (ii) \textit{reward ambiguity}: aggregating multiple criteria into a single scalar reward obscures criterion-specific token contributions and may penalize useful behaviors.
We show that these limitations lead to two types of failure criteria:

\noindent\textbullet\ \textbf{\textit{Unexplored Criteria}} correspond to behaviors absent from the current rollout group due to limited exploration (e.g., C4 in Figure \ref{fig:blind_criteria}).
Consider the setting where each criterion-wise score is binary, i.e., $r_{ij}=c_j(x,y_i)\in\{0,1\}$, indicating whether the corresponding criterion is satisfied or not. The unexplored criteria are defined as criteria that are not satisfied by any rollout in the current group:
\begin{equation}
\mathcal{C}_u = \{c_j \in \mathcal{C} \mid \forall y_i,\, r_{ij} = 0\}.
\end{equation}

\noindent\textbullet\ \textbf{\textit{Suppressed Criteria}} correspond to behaviors that the model has already discovered, but the criterion-satisfying rollouts receive non-positive aggregate advantages due to reward ambiguity (e.g., C3 in Figure \ref{fig:blind_criteria}).
Specifically, we consider the following two suppression scenarios: (i)~\emph{negative-advantage suppression}, where the advantage sum across all satisfying rollouts is negative, causing the criterion-satisfying behavior to be penalized, and (ii)~\emph{zero-advantage suppression}, where the advantage sum is zero, and the criterion is rarely satisfied (e.g., by not more than half of the group), so that the criterion receives negligible optimization pressure despite being insufficiently learned.
Let $\mathcal{S}_j = \{i \mid r_{ij} = 1\}$ denote the rollout set that satisfies criterion $c_j$, the suppressed criteria are defined as:
\begin{equation}
\mathcal{C}_s = \left\{c_j \in \mathcal{C} \;\middle|\; \mathcal{S}_j \neq \emptyset \;\wedge\; \left(\sum_{i \in \mathcal{S}_j} A_i < 0 \;\lor\; \left( \sum_{i \in \mathcal{S}_j} A_i = 0 \;\wedge\; |\mathcal{S}_j| \leq G/2 \right)\right) \right\}.
\end{equation}
Figure~\ref{fig:blind_criteria} (right) shows that both unexplored criteria and suppressed criteria are persistent and non-negligible in practice. More statistics of these two failure modes on other settings and using different LLM judges can be found in Appendix \ref{app:more_bc_statistics}.

\paragraph{On-Policy Self-Distillation (OPSD)} \citep{opsd,sdpo,opcd} has recently emerged as a promising paradigm for converting privileged information into dense token-level supervision while avoiding the training-inference distribution mismatch~\citep{opd,minillm}.
Given an on-policy rollout $y \sim \pi_{\theta}(\cdot \mid x)$, OPSD compares two
next-token distributions produced by the same model: a student distribution conditioned only on the original context, and a self-teacher distribution conditioned additionally on privileged information $\xi$. Specifically, the policy is trained to match the privilege-conditioned prediction through a token-level divergence at each token position $t$ (reverse-KL instantiation):

\begin{equation}
\mathcal{L}_{\mathrm{OPSD}}
=
\mathbb{E}_{x,y\sim\pi_\theta(\cdot|x)}\left[\frac{1}{|y|}
\sum_{t=1}^{|y|}
D_{\mathrm{KL}}
\left(
\pi_\theta(\cdot \mid x, y_{<t})
\,\middle\|\,
\operatorname{sg}\!\left(\pi_\theta(\cdot \mid x, \xi, y_{<t})\right)
\right)\right],
\label{eq:opsd}
\end{equation}
where $\operatorname{sg}(\cdot)$ denotes the stop-gradient operation. Compared with standard RLVR methods that optimize sparse outcome rewards, OPSD provides fine-grained supervision at each token position and can reveal how the privileged information changes the model's local generation preference.

In rubric-based RL, rubric criteria can naturally serve as the privileged information $\xi$ for OPSD. By conditioning the self-teacher on rubrics, OPSD can expose \textit{criterion-level supervision}, making it well suited to address the two limitations discussed above. However, directly relying on OPSD as the primary optimization objective may introduce unreliable token-level supervision due to issues such as privileged-information leakage or entropy collapse \citep{rlsd,why_sd_degrade,many_face,srpo,rlrt}. 
To this end, \textbf{we retain GRPO as the stable, reward-grounded optimization backbone and use OPSD-derived signals only for targeted criterion-level correction}, which follows the emerging paradigm of combining OPSD with RLVR rather than relying on standalone OPSD \citep{srpo,rlsd,rlrt,trace,sdpg}.

\section{\ours{}: \oursfull{}}
\label{sec:method}
To simultaneously address the two failures in rubric-based RL without introducing training-inference mismatch, we propose \textbf{\oursfull{}} (\textbf{\ours{}}).
The overall framework of \ours{} is shown in Figure~\ref{fig:overview} and the pseudo-code of \ours{} is shown in Algorithm \ref{alg:cripo}.

\subsection{Behavior Injection for Unexplored Criteria}
\label{sec:unexplored_injection}

\begin{figure}[t]
    \centering
    \includegraphics[width=1.0\linewidth]{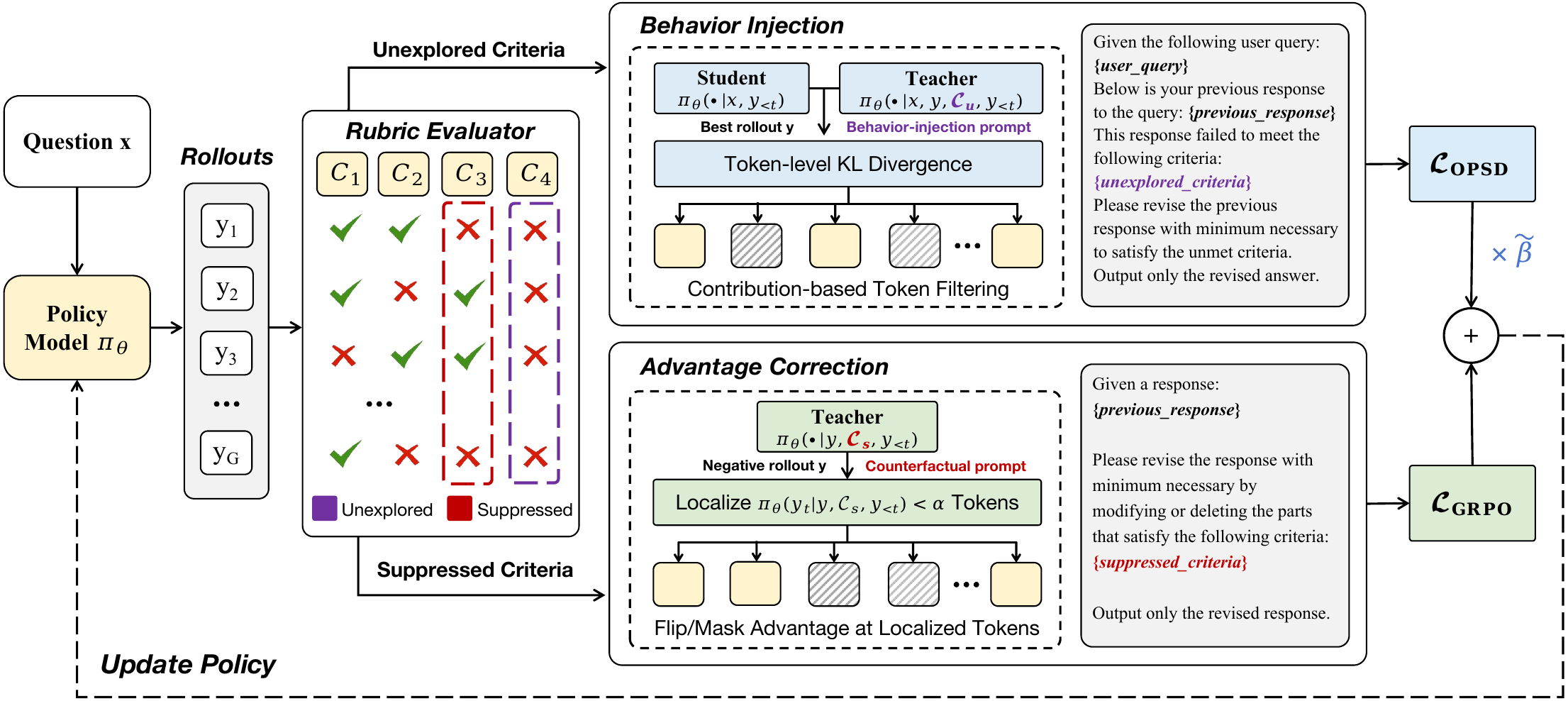}
    \caption{\textbf{Overview of \ours{}.}
\ours{} first identifies unexplored and suppressed criteria from rubric-evaluated rollouts. For unexplored criteria, \ours{} injects the missing criterion-specific behavior through \textcolor{opsdblue}{\textbf{OPSD}} Loss. For suppressed criteria, \ours{} locates criterion-relevant tokens and masks or flips their advantages to modify \textcolor{grpogreen}{\textbf{GRPO}} Loss. The two objectives are combined to update the policy.}
    \label{fig:overview}
\end{figure}

To inject the missing behaviors into the policy, we construct a \textit{behavior-injection self-teacher} to compute the auxiliary OPSD loss, which aims to revise the previous response given the unexplored criteria. The prompt template is shown in the prompt box in Figure \ref{fig:overview} at the top.
Formally, let $p^S_{t} = \pi_\theta(\cdot \mid x, y_{<t})$ and $p^T_{t} = \pi_\theta(\cdot \mid x, y, \mathcal{C}_u, y_{<t})$ denote the student's and the teacher's distributions at token position $t$, we compute the per-token \textbf{forward-KL} divergence $d_{t} = D_{\mathrm{KL}}\left(\mathrm{sg}(p^T_{t}) \,\middle\|\, p^S_{t}\right)$ to distill the teacher's criterion-conditioned behavior to the student.
We adopt forward KL since it enables the student to cover teacher-preferred modes, which encourages exploration.

\begin{wrapfigure}{r}{0.4\textwidth}
    \centering
    \vspace{-12pt}
    \includegraphics[width=0.4\textwidth]{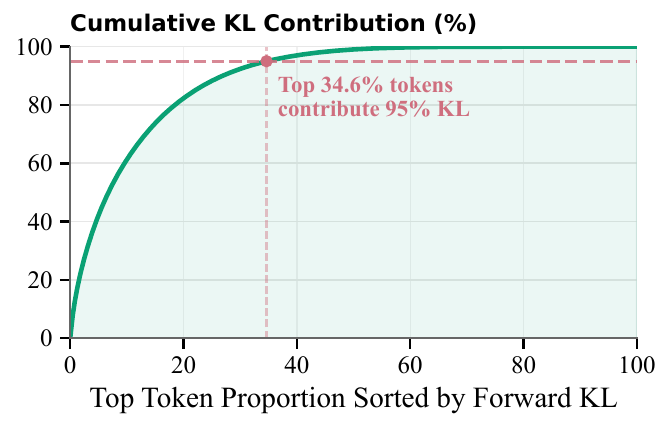}
    \vspace{-20pt}
    \caption{Cumulative KL contribution across tokens sorted by magnitude.}
    \label{fig:kl_cumulative}
    \vspace{4pt}
\end{wrapfigure}
\paragraph{Contribution-based Token Filtering.} In open-ended generations, only a few positions are actually needed to be modified to satisfy the missing criteria---the teacher distribution shifts substantially only at some tokens that affect criterion-specific signals, while the remaining tokens exhibit small KL values that primarily reflect perturbations from prompt modification. As shown in Figure~\ref{fig:kl_cumulative}, when training Qwen3-4B on the first data batch of RaR-Medicine, only 34.6\% of tokens account for 95\% of the total divergence. We therefore apply distillation only on the smallest set of tokens whose cumulative KL contribution reaches a threshold $\gamma$:
\begin{equation}
\mathcal{T}^u = \operatorname{TopCum}\left(\{d_{t}\}_{t=1}^{|y|};\, \gamma\right),
\label{eq:kl_selection}
\end{equation}
where we set $\gamma=0.95$ by default.
Moreover, for each rollout group, we only compute the distillation loss on the \textbf{highest-advantage rollout}. Since this rollout already satisfies the largest proportion of criteria, the self-teacher only needs to make minimal revisions to incorporate the remaining unexplored criteria, yielding more reliable and less noisy corrections than revising an arbitrary or low-quality response.
Let $y^*$ be the highest-advantage rollout in the group; the overall loss of behavior injection is as follows:
\begin{equation}
\mathcal{L}_{\mathrm{OPSD}}
=
\mathbb{E}_{x,\{y_i\}_{i=1}^G \sim \pi_\theta(\cdot|x)}
\left[
\frac{1}{|\mathcal{T}^u|}
\sum_{t \in \mathcal{T}^u}
D_{\mathrm{KL}}
\left(
\operatorname{sg}\!\left(\pi_\theta(\cdot \mid x, y^*, \mathcal{C}_u, y^*_{<t})\right)
\,\middle\|\,
\pi_\theta(\cdot \mid x, y^*_{<t})
\right)
\right],
\label{eq:behavior_injection}
\end{equation}

\subsection{Advantage Correction for Suppressed Criteria}
\label{sec:suppressed_flipping}

For suppressed criteria, we apply GRPO advantage correction for negative-advantage rollouts in the group, preserving the criterion-satisfying parts in them without reinforcing the entire trajectory. 

\paragraph{Suppressed-Criterion Token Localization.}
For each negative-advantage rollout $y_i$ and its satisfying suppressed criteria $\mathcal{C}_s^i=\{c_j\in\mathcal{C}_s \mid r_{ij}=1\}$, we construct a \textit{counterfactual self-teacher} that aims to revise the rollout by modifying or deleting the parts that satisfy these suppressed criteria; see the prompt template shown at the bottom of Figure~\ref{fig:overview}. Because the counterfactual teacher is explicitly instructed to remove all suppressed-criterion behaviors present in $y_i$, it assigns low probability to tokens that express those behaviors.
Let $p^T_{i,t}=\pi_\theta(\cdot\mid x,y_i,\mathcal{C}_s^i,y_{i,<t})$ denote the corresponding teacher distribution at position $t$. We identify candidate criterion-relevant tokens in rollout $y_i$ using a low-probability threshold $\alpha$:
\begin{equation}
    \mathcal{T}^s_i
    =
    \left\{t \;\middle|\; p^T_{i,t}(y_{i,t}) < \alpha\right\},
\label{eq:flip_select}
\end{equation}
The polarized distribution in Figure~\ref{fig:alpha_choice} shows that the teacher usually retains the original token with high confidence or strongly disfavors it, providing an empirical basis for using low teacher probability as a localization signal. In practice, we simply set $\alpha=0.1$ to select the low-probability subset. We also apply two cutoff strategies to these candidate tokens, as detailed in Appendix~\ref{sec:exp_setting}.

\paragraph{Selective Advantage Flipping and Masking.}
Because the localized tokens inherit the negative rollout-level advantage $A_i < 0$, GRPO penalizes them despite their contribution to satisfying suppressed criteria. A direct correction is to flip all these tokens' advantages to a small positive value $\tau_{\mathrm{flip}}$. 
However, unconditionally flipping every localized token may alter GRPO's original advantage structure more than necessary.
To this end, we use masking when the group already provides positive support for these patterns. Specifically, let $\mathcal{P}_j = \{k \mid r_{kj}=1, A_k>0\}$ denote the positive-advantage rollouts satisfying $c_j$. If $\mathcal{P}_j$ is nonempty for every $c_j\in\mathcal{C}_s^i$, we mask the localized tokens to prevent negative updates; otherwise, we flip them to provide a positive signal for criteria without such support. Let $z_i=\mathbb{I}\!\left[\exists c_j\in\mathcal{C}_s^i:\mathcal{P}_j=\emptyset\right]$ be the factor that captures this decision; the corrected token-level advantage is
\begin{equation}
    \tilde{A}_{i,t}=A_i\,\mathbb{I}\!\left[t\notin\mathcal{T}^s_i\right]+z_i\tau_{\mathrm{flip}}\,\mathbb{I}\!\left[t\in\mathcal{T}^s_i\right],
\label{eq:corrected_advantage}
\end{equation}
where we set $\tau_{\mathrm{flip}}=0.1$ by default. The corrected GRPO loss is computed using $\tilde{A}_{i,t}$ in place of the rollout-level advantage:
\begin{equation}
    \mathcal{L}_{\mathrm{GRPO}}
    =
    -\mathbb{E}_{x,\{y_i\}_{i=1}^G\sim\pi_{\theta_\mathrm{old}}(\cdot|x)}\left[\frac{1}{\sum_i|y_i|}\sum_{i=1}^{G}
    \sum_{t=1}^{|y_i|}
    \min\!\left(
    \rho_{i,t}\tilde{A}_{i,t},
    \mathrm{clip}\left(\rho_{i,t},1-\epsilon_{\mathrm{c}},1+\epsilon_{\mathrm{c}}\right)\tilde{A}_{i,t}
    \right)\right],
\label{eq:grpo_flip}
\end{equation}

\subsection{Training Objective}
\label{sec:training_objective}

To coordinate the optimization strengths of GRPO loss and OPSD loss, we adapt the OPSD coefficient to the relative-quality composition of each rollout group. Specifically, when only a few rollouts have non-negative advantages, many rollouts require direct reward-based correction, and we therefore let GRPO dominate. Conversely, when most rollouts have non-negative advantages, only a small subset requires strong negative correction from GRPO, leaving more optimization capacity for OPSD to target missing behaviors. The overall training objective is
\begin{equation}
\begin{gathered}
    \mathcal{L}_\mathrm{\ours{}}
    = \mathcal{L}_{\mathrm{GRPO}} + \tilde{\beta} \cdot \mathcal{L}_{\mathrm{OPSD}}, \\
    \text{where}\quad \tilde{\beta}
    = \left(1 + \operatorname{sign}\!\left(s-\tfrac{1}{2}\right)
    \left|2s-1\right|^{t}\right)\beta,
    \quad s=\frac{1}{G}\sum\nolimits_{i=1}^{G}\mathbb{I}\!\left[A_i\geq 0\right].
\end{gathered}
\label{eq:cripo}
\end{equation}
Here, $s$ is the proportion of rollouts with non-negative advantage in the group, $\beta$ is the base OPSD coefficient, and $t$ is a temperature parameter that controls how sensitively the coefficient responds to $s$.
We set $t=1$ in all experiments, in which case $\tilde{\beta}=2s\beta$, and during the final $25\%$ of training steps, we linearly decay the base coefficient $\beta$ to zero to stabilize late-stage optimization.

\section{Experiment}

\subsection{Experimental Setup}
\label{sec:experimental_setup}

\paragraph{Datasets and Evaluation.}
We evaluate \ours{} on both medicine and science QA tasks, where RaR-Medicine and RaR-Science \citep{rubrics_as_rewards} are adopted as the training datasets for medicine and science tasks, respectively.
We use their own test split for in-domain evaluation, and for cross-domain evaluation, we use
HealthBench~\citep{healthbench} and LLMEval-Med~\citep{llm_eval_med} for medicine tasks, and ResearchQA~\citep{researchqa} for science tasks.
We use Qwen3-32B~\citep{qwen3} as the rubric judge during training and GPT-4o-mini~\citep{gpt4o} as the judge for evaluation. For every evaluated response, the evaluation score is computed using the scalar rubric-reward aggregation defined in Section~\ref{sec:preliminary}. More details about the datasets and evaluation settings are provided in Appendix~\ref{sec:exp_setting}.

\paragraph{Baselines and Models.}
Our main experiments are conducted on Qwen3-1.7B and Qwen3-4B, and the baselines include: 
\textbf{SFT}, which fine-tunes the model to fit synthesized targets generated by rubric guidance;
\textbf{OPSD} \citep{opsd}, an on-policy self-distillation baseline that distills the full rubric into the model;
\textbf{GRPO} \citep{grpo}, the standard group-relative policy optimization baseline;
\textbf{RuscaRL} \citep{ruscarl}, a rubric-based RL method that uses rubric scaffolding during rollout generation and policy optimization;
and \textbf{HeRL} \citep{herl}, a rubric-based RL method that leverages hindsight feedback from failed trajectories and unmet rubrics to enhance exploration.
All methods are trained on the same data and evaluated under the same protocol for fair comparison. More implementation details of baselines and \ours{} are provided in Appendix \ref{sec:exp_setting}.


\subsection{Main Results}

\paragraph{\ours{} outperforms existing baselines.}
As shown in Table~\ref{tb:main_results}, SFT underperforms the base model on most benchmarks, suggesting that fitting rubric-guided targets does not generalize reliably. Standalone OPSD is also unsatisfactory without a stable reward-grounded backbone. GRPO consistently improves the base model, yielding average gains of roughly $4$--$5$ points across the two model scales, confirming the benefit of reward-grounded optimization. RuscaRL and HeRL further strengthen GRPO through rubric-guided exploration.
Built on GRPO, \ours{} raises the average score from $61.2$ to $\mathbf{64.1}$ on Qwen3-1.7B ($\mathbf{+2.9}$) and from $69.1$ to $\mathbf{72.4}$ on Qwen3-4B ($\mathbf{+3.3}$). It also achieves the best overall result, outperforming the strongest baseline by $1.1$ on Qwen3-1.7B and $1.2$ on Qwen3-4B. These results demonstrate the effectiveness of our on-policy framework that simultaneously addresses the two failure modes in standard rubric-based RL.


\begin{table*}[!t]
\caption{Experimental results ($\%$) across medicine and science benchmarks using GPT-4o-mini as judge. The best results are marked in \textbf{bold}, and the second-best are \second{underlined}. Arrows indicate changes over the base model. All results are evaluated on models trained after 400 steps.}
\vspace{0.4em}
\centering
\renewcommand{\arraystretch}{1.25}
\resizebox{\textwidth}{!}{
\begin{tabular}{l|ccc|cc|c}
\toprule
\multirow{2}{*}{\textbf{Method}}
& \multicolumn{3}{c|}{\textbf{Medicine}}
& \multicolumn{2}{c|}{\textbf{Science}}
& \multirow{2}{*}{\textbf{Avg.}} \\
\cmidrule(lr){2-4} \cmidrule(lr){5-6}
& \textbf{RaR-Medicine}
& \textbf{HealthBench}
& \textbf{LLMEval-Med}
& \textbf{RaR-Science}
& \textbf{ResearchQA}
& \\
\midrule

\textbf{Qwen3-1.7B}
& 48.6\basex{0.0} & 61.6\basex{0.0} & 56.5\basex{0.0} & 63.3\basex{0.0} & 57.2\basex{0.0} & 57.4\basex{0.0} \\
\hspace*{1em}+ SFT
& 45.8\down{2.8} & 50.1\down{11.5} & 46.5\down{10.0} & 62.1\down{1.2} & 47.0\down{10.2} & 50.3\down{7.1} \\
\hspace*{1em}+ OPSD
& 48.5\down{0.1} & 51.9\down{9.7} & 54.3\down{2.2} & 64.8\uprise{1.5} & 52.5\down{4.7} & 54.4\down{3.0} \\
\hspace*{1em}+ GRPO
& 50.7\uprise{2.1} & 63.0\uprise{1.4} & 60.6\uprise{4.1} & 68.3\uprise{5.0} & 63.6\uprise{6.4} & 61.2\uprise{3.8} \\
\hspace*{1em}+ RuscaRL
& 51.5\uprise{2.9} & 63.8\uprise{2.2} & \second{62.1}\uprise{5.6} & \second{70.7}\uprise{7.4} & \second{64.8}\uprise{7.6} & 62.6\uprise{5.2} \\
\hspace*{1em}+ HeRL
& \second{52.6}\uprise{4.0} & \second{65.8}\uprise{4.2} & \best{64.2}\uprise{7.7} & 70.1\uprise{6.8} & 62.3\uprise{5.1} & \second{63.0}\uprise{5.6} \\
\rowcolor{darkrowgreen} \hspace*{1em}+ \textbf{\ours{}}
& \best{52.8}\uprise{4.2} & \best{67.3}\uprise{5.7} & \best{64.2}\uprise{7.7} & \best{71.0}\uprise{7.7} & \best{65.0}\uprise{7.8} & \best{64.1}\uprise{6.7} \\

\midrule

\textbf{Qwen3-4B}
& 55.8\basex{0.0} & 65.7\basex{0.0} & 68.4\basex{0.0} & 69.2\basex{0.0} & 62.1\basex{0.0} & 64.2\basex{0.0} \\
\hspace*{1em}+ SFT
& 54.6\down{1.2} & 55.2\down{10.5} & 59.2\down{9.2} & 71.1\uprise{1.9} & 54.0\down{8.1} & 58.8\down{5.4} \\
\hspace*{1em}+ OPSD
& 56.6\uprise{0.8} & 55.8\down{9.9} & 58.5\down{9.9} & 71.1\uprise{1.9} & 58.5\down{3.6} & 60.1\down{4.1} \\
\hspace*{1em}+ GRPO
& 60.0\uprise{4.2} & 69.8\uprise{4.1} & 69.9\uprise{1.5} & 76.0\uprise{6.8} & 70.0\uprise{7.9} & 69.1\uprise{4.9} \\
\hspace*{1em}+ RuscaRL
& \second{63.2}\uprise{7.4} & 72.1\uprise{6.4} & 71.4\uprise{3.0} & \second{77.1}\uprise{7.9} & \second{72.4}\uprise{10.3} & \second{71.2}\uprise{7.0} \\
\hspace*{1em}+ HeRL
& 62.4\uprise{6.6} & \best{73.2}\uprise{7.5} & \second{71.7}\uprise{3.3} & 74.5\uprise{5.3} & 71.7\uprise{9.6} & 70.7\uprise{6.5} \\
\rowcolor{darkrowgreen} \hspace*{1em}+ \textbf{\ours{}}
& \best{63.5}\uprise{7.7} & \second{72.6}\uprise{6.9} & \best{73.9}\uprise{5.5} & \best{78.6}\uprise{9.4} & \best{73.6}\uprise{11.5} & \best{72.4}\uprise{8.2} \\

\bottomrule
\end{tabular}}
\label{tb:main_results}
\end{table*}

\begin{figure}[t]
    \centering
    \includegraphics[width=1\linewidth]{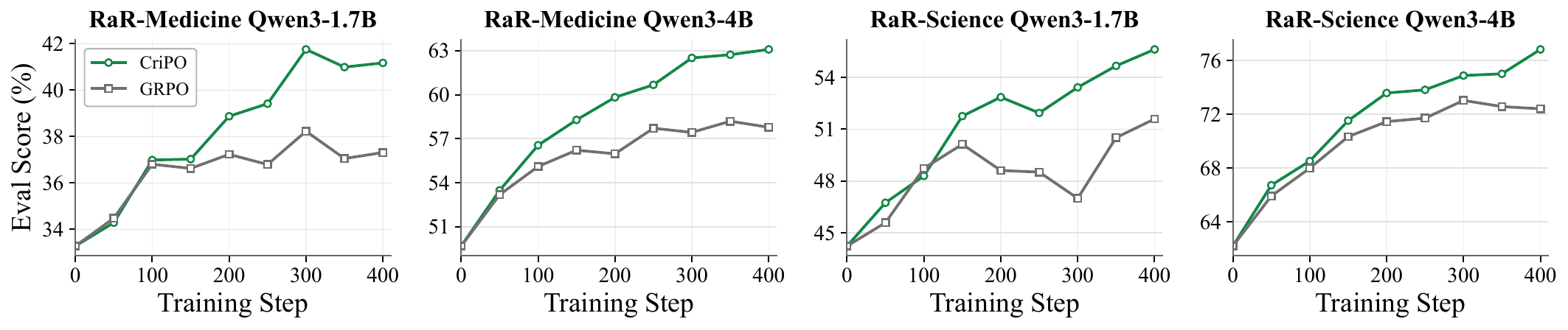}
    \caption{\textbf{Comparisons of evaluation scores during training.} Scores are judged by Qwen3-32B.}
    \label{fig:compute_efficiency}
\end{figure}

\paragraph{Efficiency and Convergence Analysis.}
Figure~\ref{fig:compute_efficiency} compares the optimization trajectories of \ours{} and GRPO over 400 training steps. GRPO improves rapidly in the early stage, but its subsequent gains become slower and less consistent, whereas \ours{} exhibits more sustained improvement. 
Across all model-domain settings, \ours{} reaches or exceeds the best score attained by GRPO within the first 150 to 200 steps and continues toward a substantially higher final score, demonstrating more efficient use of training samples and policy updates.
In terms of the training time, \ours{} requires at most one additional forward pass per rollout, resulting in about a 7\% increase on average in training time compared with GRPO.
Detailed comparisons of wall-clock training time as well as peak GPU memory are provided in Appendix~\ref{app:compute_cost}.

\subsection{Analysis}

\begin{figure}[t]
    \centering
    \includegraphics[width=1\linewidth]{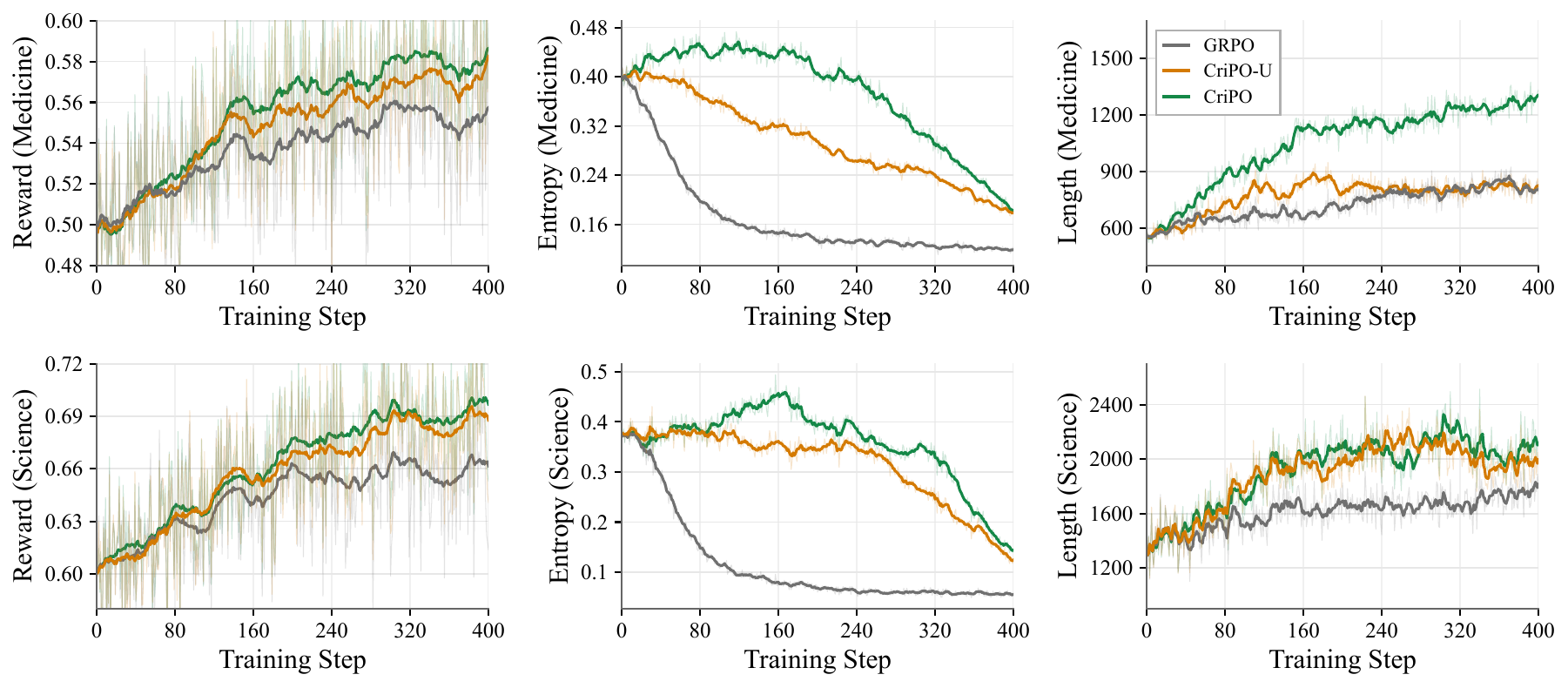}
    \caption{\textbf{Training dynamics of \ours{}.}
We compare \ours{} and GRPO across reward, entropy, and response length over 400 training steps when training Qwen3-4B on RaR-Medicine and RaR-Science.}
    \label{fig:dynamics}
\end{figure}

\paragraph{Training Dynamics.}
To examine how the two modules in \ours{} affect optimization, Figure~\ref{fig:dynamics} compares GRPO, \ours{}-U (behavior injection only), and \ours{} over 400 steps of training Qwen3-4B on RaR-Medicine and RaR-Science.
As shown in the \textbf{left} column, \ours{}-U consistently follows a higher reward trajectory than GRPO, while adding advantage correction generally yields the highest reward, demonstrating the complementary benefits of the two modules.
The \textbf{middle} column shows that GRPO undergoes rapid entropy collapse, whereas behavior injection substantially slows this decline and thereby preserves exploration. The full \ours{} maintains even higher entropy through most of training, suggesting that advantage correction further protects diverse criterion-relevant patterns from negative updates. As the base OPSD coefficient $\beta$ is linearly decayed toward zero in the final training stage, the entropy of both \ours{} variants also decreases, allowing the policies to settle into more stable output distributions.
The \textbf{right} column further shows that \ours{} produces longer responses than GRPO in both domains. Behavior injection encourages the model to cover missing criterion-specific behaviors, while advantage correction preserves content associated with suppressed criteria; the latter produces a particularly clear additional length increase on RaR-Medicine. Together with the reward gains, these trends suggest that the increased length reflects richer rubric-relevant content rather than mere verbosity. Appendix~\ref{app:case_study} provides an illustrative example in which \ours{} gives more complete clinical reasoning and satisfies more rubric criteria than GRPO.

\begin{figure}[t]
    \centering
    \includegraphics[width=1\linewidth]{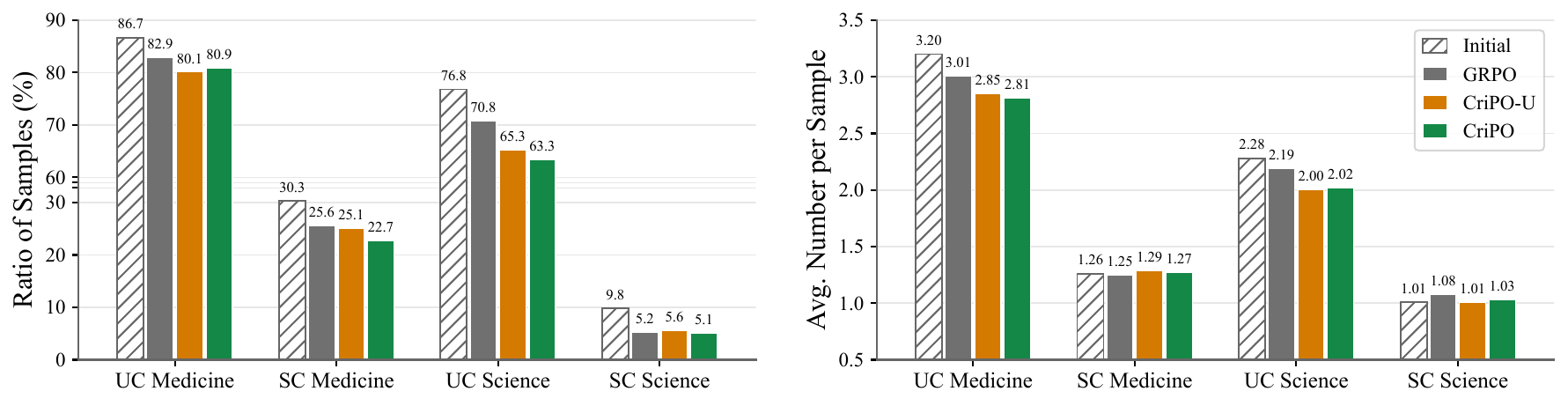}
    \caption{\textbf{Statistics of unexplored and suppressed criteria on the testing set.}
We report the occurrence ratio and average count across methods trained with Qwen3-4B.}
    \label{fig:bc_val}
\end{figure}

\paragraph{Statistics of Unexplored and Suppressed Criteria.}
To examine whether the two modules alleviate their targeted failure modes, Figure~\ref{fig:bc_val} reports the statistics on the RaR-Medicine and RaR-Science testing sets for methods trained with Qwen3-4B. We use the training-time sampling configuration rather than that of evaluation to better approximate the training rollout distribution.
The following two observations emerge. First, \textbf{behavior injection substantially reduces UC}. Compared with the initial model, GRPO lowers the UC ratio by $3.8$ and $6.0$ percentage points on medicine and science. Adding behavior injection (\ours{}-U) yields further reductions of $2.8$ and $5.5$ points over GRPO, while decreasing the average UC count from $3.01$ to $2.85$ on medicine and from $2.19$ to $2.00$ on science. These improvements show that behavior injection helps the policy acquire criterion-specific behaviors absent from its rollout groups.
Second, \textbf{advantage correction further mitigates SC}. Relative to \ours{}-U, the full \ours{} reduces the SC ratio from $25.1\%$ to $22.7\%$ on medicine and from $5.6\%$ to $5.1\%$ on science, and the average SC count remains nearly unchanged on both domains. Also, adding advantage correction largely preserves the UC gains and even reduces the UC ratio on science, consistent with the intuition that preserving suppressed behaviors prevents explored criteria from becoming unexplored in subsequent rollouts.
Overall, each module primarily improves its intended failure mode, while their combination provides the most balanced reduction of UC and SC.

\begin{wraptable}{r}{0.45\textwidth}
\vspace{-1.8em}
\caption{Comparisons between the behavior-injection teacher rollouts and the student rollouts on the testing set.}
\vspace{0.3em}
\centering
\scriptsize
\setlength{\tabcolsep}{2.5pt}
\renewcommand{\arraystretch}{1.10}
\renewcommand{\uprise}[1]{\textcolor{mygreen}{\tiny\;{$\uparrow$ #1}}}
\resizebox{\linewidth}{!}{
\begin{tabular}{lcc|cc}
\toprule
\multirow{2}{*}{\textbf{Setting}}
& \multicolumn{2}{c|}{\textbf{Rollout Reward}}
& \multicolumn{2}{c}{\textbf{UC Number}} \\
\cmidrule(lr){2-3} \cmidrule(lr){4-5}
& \textbf{Student} & \textbf{Teacher}
& \textbf{Student} & \textbf{Recovered} \\
\midrule
Qwen3-1.7B Medicine & 53.0 & 73.0\uprise{20.0} & 3.86 & 1.93 \\
Qwen3-4B Medicine   & 60.7 & 75.1\uprise{14.4} & 3.20 & 1.98 \\
Qwen3-1.7B Science  & 58.5 & 74.5\uprise{16.0} & 2.92 & 1.25 \\
Qwen3-4B Science    & 67.0 & 82.3\uprise{15.3} & 2.29 & 1.46 \\
\bottomrule
\end{tabular}}
\label{tab:teacher_reliability}
\vspace{-1.5em}
\end{wraptable}

\paragraph{Reliability of Behavior-Injection Teacher.}
To assess whether the behavior-injection teacher provides reliable supervision, we directly sample four teacher rollouts for samples and use the highest-reward rollout as a proxy for the teacher's ability to elicit the missing behaviors. Table~\ref{tab:teacher_reliability} compares the rewards of student and teacher rollouts and reports the student's average UC count. "Recovered" denotes the number of the student's UC that are satisfied by the selected teacher rollout.
Across all model-domain settings, the teacher consistently produces higher-reward responses than the student, with gains of $14.4$--$20.0$ points. It also recovers $1.25$--$1.98$ UC per sample, corresponding to approximately $43$--$64\%$ of the criteria missed by the student. These consistent improvements support distilling the teacher's token-level signal into the student for behavior injection.

\paragraph{What Tokens Does the Counterfactual Teacher Locate?}
\begin{table}[t]
\caption{\textbf{Types of tokens localized by the counterfactual teacher.}
Statistics are computed over 8,497 selected tokens from 1,641 suppressed-criterion rollout assignments.}
\vspace{0.3em}
\centering
\scriptsize
\setlength{\tabcolsep}{5pt}
\renewcommand{\arraystretch}{1.05}
\resizebox{\linewidth}{!}{
\begin{tabular}{p{0.25\textwidth}p{0.55\textwidth}c}
\toprule
\textbf{Token Category} & \textbf{Representative Tokens} & \textbf{Proportion} \\
\midrule
Formatting \& Structural Boundaries & \texttt{\textbackslash n}, \texttt{\textbackslash n\textbackslash n}, \texttt{.}, \texttt{:}, \texttt{(}, \texttt{**}, \texttt{\#\#\#}, \texttt{---}, \texttt{-} & 44.3\% \\
Knowledge \& Semantic Content & \texttt{clinical}, \texttt{blood}, \texttt{therapy}, \texttt{liver}, \texttt{Fever}, \texttt{virus} & 21.5\% \\
Logical \& Discourse Connectors & \texttt{and}, \texttt{that}, \texttt{which}, \texttt{However}, \texttt{Let}, \texttt{Other}, \texttt{Explanation}, \texttt{If}, \texttt{but} & 17.1\% \\
Grammatical Function Words & \texttt{the}, \texttt{a}, \texttt{in}, \texttt{of}, \texttt{to}, \texttt{is}, \texttt{this} & 15.4\% \\
Numbers \& Option Labels & \texttt{1}, \texttt{2}, \texttt{3}, \texttt{4}, \texttt{A}, \texttt{B}, \texttt{C}, \texttt{D} & 1.6\% \\
\bottomrule
\end{tabular}}
\label{tab:localized_token_types}
\end{table}

We analyze the tokens localized by the counterfactual teacher using Qwen3-4B on the RaR-Medicine testing set, and their major categories and distributions are summarized in Table~\ref{tab:localized_token_types}. Advantage correction on formatting and structural boundaries can preserve the organization and extent of criterion-bearing spans, correction on logical and discourse connectors can protect useful reasoning relations and points of elaboration, correction on knowledge and semantic content can retain criterion-relevant facts and entities, and correction on numbers and option labels can preserve quantitative information and final-answer choices. In contrast, grammatical function words carry little meaning in isolation and may provide limited value for advantage correction. Appendix~\ref{app:suppressed_token_cases} provides a criterion-wise breakdown and representative success and failure cases.

Please refer to more results and analysis in the Appendix, including results using Qwen3-32B as judge (\ref{app:judge_robustness}), other LLMs (\ref{app:other_models}), multiple seeds (\ref{app:seeds}), different OPSD variants (\ref{app:opsd_variants}), OOD evaluation (\ref{app:ood}), using DAPO as the optimization backbone (\ref{app:dapo}), and ablation studies (\ref{app:ablation}).

\section{Conclusion}

In this work, we study rubric-based RL for improving LLMs on open-ended tasks and identify two prevalent criterion-level failure modes: unexplored criteria, where no rollout satisfies a criterion, and suppressed criteria, where criterion-satisfying behaviors are penalized or ignored due to non-positive aggregate advantages. To address both issues, we propose \oursfull{} (\ours{}), an on-policy framework that retains GRPO as a reward-grounded backbone while using OPSD for targeted criterion-level correction. Specifically, \ours{} injects unexplored behavior through a localized forward-KL distillation loss and preserves useful suppressed behaviors by locating relevant tokens and correcting their advantages. Experiments on medicine and science benchmarks demonstrate the effectiveness and efficiency of \ours{} against existing rubric-based RL methods. We hope our work can inspire the community to further investigate how to jointly address both limited exploration and reward ambiguity in standard rubric-based RL.

\section*{AI use statement}
In this work, we used generative AI tools to assist with translation and implement methods. We did not use generative AI tools to design or provide feedback on the research methodology or experiments, support qualitative and thematic data analysis, or interpret results. Generating synthetic datasets, developing theoretical models or conceptual frameworks, formulating mathematical claims, providing critical ingredients for mathematical proofs, assisting in writing proofs, proposing or refining hypotheses, and cleaning or reformatting datasets were not applicable to this work. Additionally, we used generative AI tools to create or edit software code, search for information, and edit the paper to improve readability. We reviewed all AI-assisted work by manually checking the AI-generated text, code, and search results. We take responsibility for the final content of this work, including text, claims, and artifacts produced with the aid of generative AI.

\bibliographystyle{iclr2027_conference}
\bibliography{main}

\clearpage
\appendix

\section{Related Work}\label{sec:related_work}

\paragraph{Rubric-based RL.}
Rubric-based RL extends RLVR to open-ended tasks whose quality cannot be verified by exact answers or executable tests. By decomposing response quality into explicit dimensions such as factuality, completeness, safety, evidence grounding, and task utility, they provide more interpretable and controllable supervision than holistic judgments~\citep{healthbench, deepresearch}. Existing rubric-based RL methods typically use an LLM judge to score individual criteria and aggregate these scores into a scalar reward for GRPO~\citep{rubrics_as_rewards, checklists, writingbench, qalign}. Subsequent work improves this paradigm through better reward aggregation, criterion reweighting, and fine-grained credit assignment, making rubric signal more discriminative and robust~\citep{rucl, alternating, rtt, cort}. A parallel line of work uses rubric feedback to enhance exploration. RuscaRL~\citep{ruscarl} conditions rollout generation on rubric guidance, while HeRL~\citep{herl} uses hindsight feedback from failed trajectories to construct revised rollouts. Although these methods help discover behaviors associated with previously unmet criteria, they rely on privileged information during rollout generation, introducing a training-inference mismatch, and also overlook another crucial failure of suppressed criteria. In contrast, \ours{} recovers criterion-level supervision for both failure modes through an on-policy framework, without relying on externally guided rollouts.

\paragraph{On-Policy Self-Distillation.}
On-Policy Self-Distillation (OPSD) uses the current policy under privileged or feedback-augmented contexts as a self-teacher, converting auxiliary information into dense token-level supervision while avoiding off-policy teacher mismatch~\citep{opd, opsd, sdpo}. Despite its promise, dense OPSD is often unstable due to noisy token-level gradients, teacher--student inconsistency, and privileged-information leakage~\citep{rlsd,why_sd_degrade,many_face,srpo,rlrt}. Recent work therefore increasingly combines OPSD with RLVR, using reference solutions, verified traces, or environment feedback to improve sparse-reward optimization~\citep{rlsd, sdar,rlcsd}. Most of these studies focus on verifiable domains such as mathematical reasoning, where privileged supervision is relatively well aligned with the objective.

Recent works have also begun exploring OPSD for rubric-based post-training in open-ended tasks~\citep{rgsd,rcsd}, while they treat self-distillation as the primary optimization objective rather than retaining a reward-grounded RL backbone. 
Concurrent work~\citep{papergym} combines OPSD and GRPO sequentially, using OPSD before RL to obtain a more knowledgeable initialization.
In contrast, \ours{} integrates self-distillation directly into GRPO training and uses it only for targeted criterion-level correction.

\section{Detailed Experimental Settings}\label{sec:exp_setting}

\paragraph{Dataset Details.}
For RaR-Medicine and RaR-Science~\citep{rubrics_as_rewards}, we first filter out overly easy samples to improve data efficiency. Specifically, we remove samples on which Qwen3-4B-Instruct achieves a reward higher than 0.9. This results in 15,658 training samples and 1,936 testing samples for RaR-Medicine, and 10,874 training samples and 1,365 testing samples for RaR-Science.
The weights of criteria are all positive for the two training datasets where pitfall criteria are phrased in positive form following \cite{rubrics_as_rewards}.
For HealthBench~\citep{healthbench} and ResearchQA~\citep{researchqa}, we randomly select a subset of 500 samples from each dataset for cross-domain evaluation. For LLMEval-Med, we follow the evaluation protocol of the original project~\citep{llm_eval_med}, which contains 667 medical questions across five categories.

\paragraph{Evaluation Details.}
During training, we use Qwen3-32B as the rubric judge, and we use GPT-4o-mini as the rubric judge in our main evaluation. The sampling temperature for the judge model is set as 0.1 for low uncertainty, and we disable the thinking mode for Qwen3-32B. The judge prompt template is shown in Figure~\ref{fig:judge_prompt}.
For evaluation, we set the policy model's generation parameters to rollout\_n$=1$, top\_k$=-1$, top\_p$=0.8$, and temperature$=0.7$ across all benchmarks.

\paragraph{Implementation Details.}
The training is implemented based on the verl \citep{verl} framework, and the OPSD implementation follows SDPO \citep{sdpo}. 
Hardware configuration details are withheld due to organizational confidentiality restrictions.
Detailed training configurations of \ours{} and all baselines are provided in Table~\ref{tab:training_config}.
We use Qwen3-32B to synthesize the training target for SFT, which is guided by the full rubric. The target generation prompt for SFT and the teacher prompt for OPSD are shown in Figure \ref{fig:full_rubric_teacher_prompt}. The KL divergence in the OPSD loss is implemented as a top-$k$ vocabulary approximation, where we use $k=64$. 

\begin{table}[ht]
\vspace{-1.5em}
\caption{Training configurations across different methods and model backbones.}
\vspace{0.2em}
\centering
\renewcommand{\arraystretch}{1.18}

\resizebox{0.95\linewidth}{!}{
\begin{tabular}{p{0.18\linewidth} p{0.72\linewidth}}
\toprule
\textbf{Settings} & \textbf{Hyperparameters} \\
\midrule
\textbf{SFT} 
& n\_gpus $=8$, train\_batch\_size $=64$, micro\_batch\_size\_per\_gpu $= 1$ \\
& learning\_rate $= 1\mathrm{e}{-5}$, lr\_scheduler\_type $=$ cosine \\
& max\_length $=8192$, warmup\_steps $=40$, total\_training\_steps $= 400$ \\
\midrule
\textbf{RL Sampling} 
& rollout\_n $= 8$, top\_k $= -1$, top\_p $= 1.0$, temperature $= 1.0$ \\
& max\_prompt\_length $= 1024$, max\_response\_length $= 5120$ \\
\midrule
\textbf{RL Training} 
& n\_gpus $=8$, train\_batch\_size $=64$, ppo\_mini\_batch\_size $= 32$ \\
& ppo\_micro\_batch\_size\_per\_gpu $= 2/1$ for RaR-Medicine/RaR-Science \\
& learning\_rate $= 1\mathrm{e}{-6}$, kl\_loss\_coef $= 1\mathrm{e}{-3}$, total\_training\_steps $= 400$ \\
\midrule
\textbf{RuscaRL/HeRL} 
& The hyperparameters in RuscaRL/HeRL follow their default settings. \\
\midrule
\textbf{OPSD}
& rollout\_n$=1$, top\_k$=-1$, top\_p$=0.8$, temperature$=0.7$ \\
& loss type: forward KL, privileged information: full rubric \\
\midrule
\textbf{\ours{}} 
& $\gamma=0.95$, $\alpha=0.1$, $\tau_\mathrm{flip}=0.1$, $\beta=0.05$, $\kappa=2048$, $\zeta=10$ \\
\bottomrule
\end{tabular}}
\label{tab:training_config}
\end{table}

\paragraph{More Implementation Details for \ours{}.}
To keep advantage correction targeted, we use two cutoffs. Let $\kappa=2048$ be the maximum response length eligible for advantage correction and $\zeta=10$ the maximum number of key tokens selected per rollout. If $|y_i|>\kappa$, we leave its token advantages unchanged. Otherwise, among the tokens satisfying the teacher-probability threshold in Eq.(\ref{eq:flip_select}), we retain at most the $\zeta$ tokens with the lowest teacher-assigned probabilities. Advantage correction can increase response length by preserving or reinforcing selected tokens; correcting too many tokens could amplify this effect and destabilize training. Longer responses also make criterion-specific token localization less reliable, motivating the response-length cutoff. Together, these cutoffs constrain the scope of advantage correction while retaining its most salient token-level signals.

\begin{figure}[ht]
\centering
\includegraphics[width=0.45\textwidth]{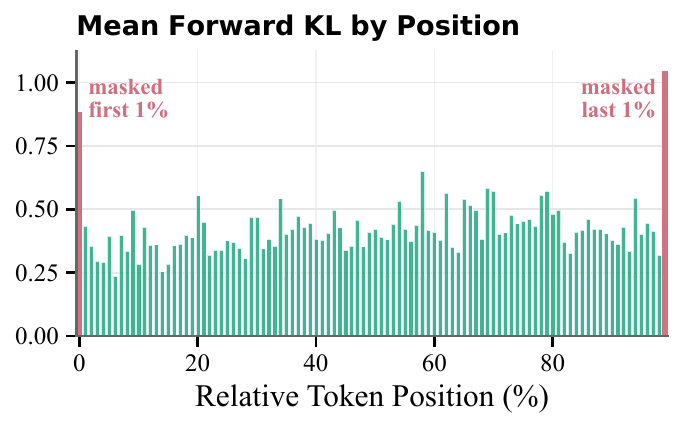}
\vspace{-0.5em}
\caption{\textbf{Token-wise forward KL across positions.}
Divergence is higher near response boundaries.}
    \label{fig:kl_position}
\end{figure}

We apply further safeguards to the behavior-injection self-teacher. We retain at most the three highest-weighted unexplored criteria when constructing the teacher prompt. We also clamp each token's KL value to a maximum of $10$ before contribution-based filtering and gradient updates to limit outliers. Finally, we observe larger KL values near the beginning and end of responses (Figure~\ref{fig:kl_position}). Early tokens may reflect the teacher instruction, such as a preamble before revising the response. At the end, the teacher is conditioned on a suboptimal student prefix and may favor post-hoc corrections rather than reliable continuation. We therefore mask the first and last $1\%$ of response tokens before contribution-based filtering. Overall, \textbf{we use the same settings for the above parameters and $\gamma$, $\alpha$, $\tau_{\mathrm{flip}}$, $\beta$, $\kappa$, and $\zeta$ across all model and dataset settings, without per-setting tuning}; see Appendix~\ref{app:ablation} for a detailed hyperparameter analysis.

\section{Supplementary Experiments}

\begin{figure}[t]
    \centering
    \includegraphics[width=1\linewidth]{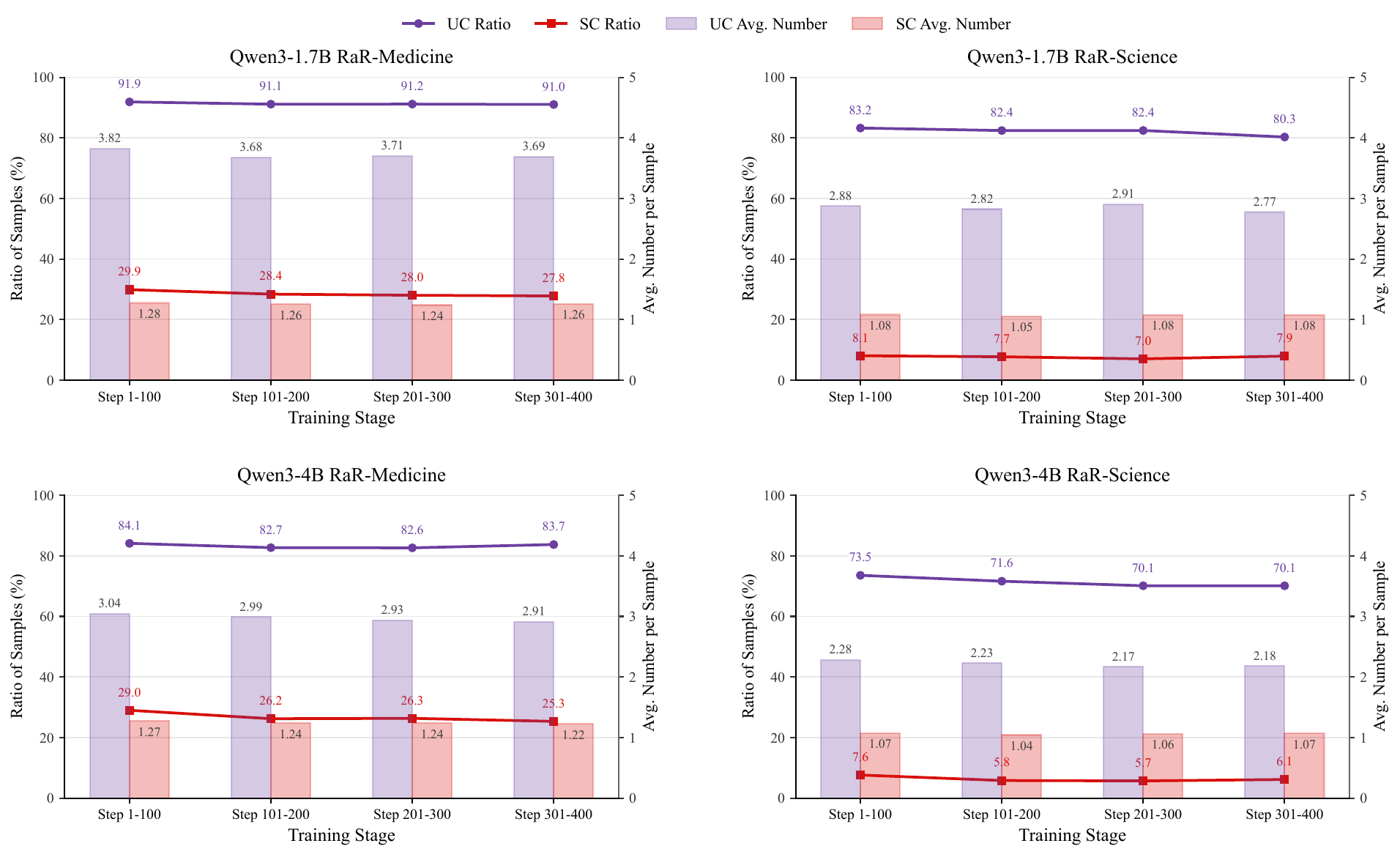}
    \caption{Statistics of unexplored and suppressed criteria of GRPO on different settings.}
    \label{fig:bc_statistics_all}
\end{figure}

\begin{table}[t]
\caption{Statistics of unexplored and suppressed criteria obtained with different judge models.}
\vspace{0.4em}
\centering
\renewcommand{\arraystretch}{1.20}
\resizebox{0.6\linewidth}{!}{
\begin{tabular}{lcccc}
\toprule
\textbf{Judge Model} & \textbf{UC Ratio} & \textbf{SC Ratio} & \textbf{UC Count} & \textbf{SC Count} \\
\midrule
Qwen3-32B    & 0.81 & 0.31 & 3.10 & 1.40 \\
GPT-4o-mini  & 0.75 & 0.25 & 2.90 & 1.19 \\
GPT-5.6-luna & 0.88 & 0.34 & 3.39 & 1.32 \\
\bottomrule
\end{tabular}}
\label{tab:judge_uc_sc}
\end{table}

\subsection{Statistics of Unexplored and Suppressed Criteria}\label{app:more_bc_statistics}

\paragraph{Statistics on Different Settings.}
We further examine UC and SC throughout GRPO training across two model scales and two domains. Figure~\ref{fig:bc_statistics_all} groups the 400 training steps into four stages and reports both the occurrence ratio and average count of each failure mode. UC remains widespread in every setting: its ratio ranges from $70.1\%$ to $91.9\%$, with an average count of $2.17$--$3.82$. Scaling from Qwen3-1.7B to Qwen3-4B generally lowers UC, particularly on RaR-Science, but does not eliminate it. SC is less frequent and more domain-dependent, with ratios of $25.3$--$29.9\%$ on RaR-Medicine and $5.7$--$8.1\%$ on RaR-Science, while its average count remains between $1.04$ and $1.28$.
Both failure modes decrease only modestly as training proceeds. For example, on Qwen3-4B, the SC ratio on RaR-Medicine decreases from $29.0\%$ in steps 1--100 to $25.3\%$ in steps 301--400, while the UC ratio on RaR-Science decreases from $73.5\%$ to $70.1\%$. In other settings, the statistics are nearly flat; the UC ratio of Qwen3-1.7B on RaR-Medicine remains around $91\%$ throughout training. These results show that standard GRPO alone does not resolve either failure mode: UC and SC persist across model scales, domains, and training stages, motivating explicit criterion-level correction.

\paragraph{Statistics using Different LLM Judges.}
To determine whether these observations depend on the choice of LLM judge, we use GPT-4o-mini and GPT-5.6-luna, in addition to the Qwen3-32B judge used during training, to independently evaluate the same Qwen3-4B rollouts for the first 64 samples of the RaR-Medicine testing set. As shown in Table~\ref{tab:judge_uc_sc}, all three judges identify UC in the majority of samples: the UC ratio ranges from $75\%$ to $88\%$, with an average count of $2.90$--$3.39$. They also consistently detect SC, with occurrence ratios of $25\%$--$34\%$ and average counts of $1.19$--$1.40$. Although the absolute estimates vary---GPT-4o-mini is the most conservative, whereas GPT-5.6-luna reports the highest UC and SC ratios---the qualitative conclusion is consistent across judges. Thus, the prevalence of both failure modes is not an artifact of the particular judge used during training.

\begin{figure}[t]
    \centering
    \includegraphics[width=1\linewidth]{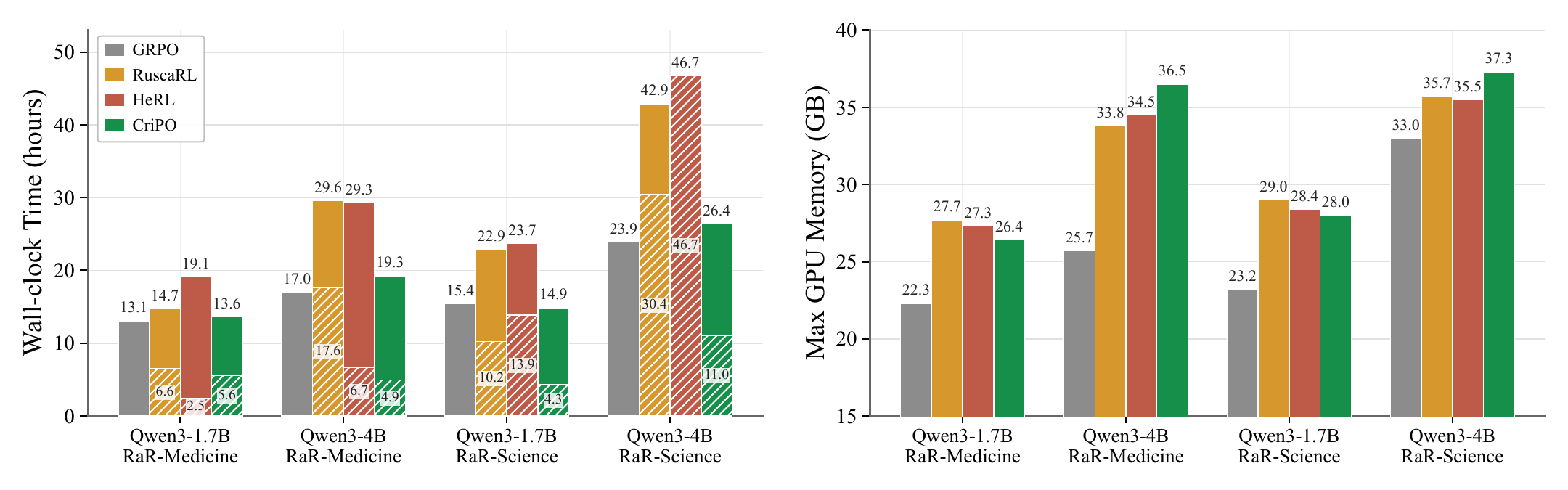}
    \caption{\textbf{Computation cost comparison.}
    \textbf{Left:} Total wall-clock training time of GRPO, RuscaRL, HeRL, and \ours{} across model scales and training domains. Numbers above the bars denote total training time, while the hatched regions and their labels indicate the time required to reach the final performance achieved by GRPO; GRPO itself serves as the reference. \textbf{Right:} Maximum per-GPU memory usage under the same settings.}
    \label{fig:cost}
\end{figure}

\subsection{Computation Cost Analysis}\label{app:compute_cost}
\paragraph{Comparisons on Training Time.} Figure~\ref{fig:cost} compares the wall-clock training time and maximum GPU memory of \ours{} with GRPO and rubric-guided baselines. With at most one additional forward pass per rollout, the total training time of \ours{} remains close to that of GRPO: Averaged over the four settings, \ours{} requires $18.6$ hours, only about $7\%$ more than GRPO ($17.4$ hours), while being substantially faster than RuscaRL ($27.5$ hours) and HeRL ($29.7$ hours).
The runtime differences arise from the distinct training pipelines: RuscaRL assigns a large weight to rubric-guided generation during the early training stage, which substantially increases both rollout entropy and response length. Because autoregressive generation cost grows with sequence length, these longer rollouts lead to higher wall-clock time. HeRL incurs a different source of overhead: each sample passes through two rollout stages and requires two calls to the LLM judge. Switching between rollout modes and performing the additional judge evaluation both add substantial latency. In contrast, \ours{} reuses the original on-policy rollout and primarily adds one extra actor forward pass to obtain the self-teacher signal, explaining why its total runtime remains much closer to GRPO.

More importantly, the faster convergence of \ours{} offsets its additional per-step cost. As indicated by the hatched regions in Figure~\ref{fig:cost} (left), \ours{} reaches GRPO's final performance in only $5.6$, $4.9$, $4.3$, and $11.0$ hours across the four settings. These values correspond to a $2.2\times$--$3.6\times$ wall-clock speedup over the time required to fully train GRPO. \ours{} is also the fastest non-GRPO method to reach the GRPO reference performance in three of the four settings; the exception is Qwen3-1.7B on RaR-Medicine, where HeRL reaches it in $2.5$ hours. The above computation cost comparison shows that \ours{} provides a favorable time-to-quality tradeoff rather than merely improving final performance through a larger computation budget.

\paragraph{Comparisons on GPU Memory.} Figure~\ref{fig:cost} (right) shows that this efficiency comes with a moderate memory overhead. The maximum per-GPU memory of \ours{} is $26.4$, $36.5$, $28.0$, and $37.3$ GB, compared with $22.3$, $25.7$, $23.2$, and $33.0$ GB for GRPO. This increase mainly arises from the top-$k$ vocabulary approximation used to compute forward KL, which requires retaining probabilities or logits for the selected vocabulary entries during the actor forward pass. Nevertheless, peak memory remains below $40$ GB in every setting and is generally comparable to RuscaRL and HeRL. Overall, \ours{} introduces manageable memory overhead while delivering substantially faster convergence and stronger final performance.

\subsection{Results using Different LLM Judges}
\label{app:judge_robustness}

The main results in Table~\ref{tb:main_results} use GPT-4o-mini as the evaluation judge. To examine whether the conclusions depend on this evaluator, we further evaluate the same set of trained models using Qwen3-32B. As shown in Table~\ref{tb:qwen3_judge_results}, the results lead to the same overall conclusions as those obtained with GPT-4o-mini: SFT and standalone OPSD are less reliable, reward-grounded RL improves the base models, and \ours{} achieves the strongest average performance at both model scales.

Specifically, SFT and standalone OPSD underperform the corresponding base models on average. GRPO raises the average score from $41.1$ to $45.3$ on Qwen3-1.7B ($+4.2$) and from $53.4$ to $60.0$ on Qwen3-4B ($+6.6$), confirming the benefit of reward-grounded optimization. RuscaRL and HeRL provide further improvements through rubric-guided exploration. Built on GRPO, \ours{} further increases the average score to $\mathbf{48.3}$ on Qwen3-1.7B ($\mathbf{+3.0}$ over GRPO) and $\mathbf{63.6}$ on Qwen3-4B ($\mathbf{+3.6}$). It also outperforms the strongest competing baseline by $1.3$ and $1.1$ points, respectively. The consistent advantage of \ours{} under both GPT-4o-mini and Qwen3-32B indicates that the benefits of targeted criterion-level correction are robust to the choice of evaluation judge.

\begin{table*}[!t]
\caption{Experimental results ($\%$) across medicine and science benchmarks using Qwen3-32B as judge. The best results are marked in \textbf{bold}, and the second-best are \second{underlined}. Arrows indicate changes over the base model. All results are evaluated on models trained after 400 steps.}
\vspace{0.4em}
\centering
\renewcommand{\arraystretch}{1.25}
\resizebox{\textwidth}{!}{
\begin{tabular}{l|ccc|cc|c}
\toprule
\multirow{2}{*}{\textbf{Method}}
& \multicolumn{3}{c|}{\textbf{Medicine}}
& \multicolumn{2}{c|}{\textbf{Science}}
& \multirow{2}{*}{\textbf{Avg.}} \\
\cmidrule(lr){2-4} \cmidrule(lr){5-6}
& \textbf{RaR-Medicine}
& \textbf{HealthBench}
& \textbf{LLMEval-Med}
& \textbf{RaR-Science}
& \textbf{ResearchQA}
& \\
\midrule

\textbf{Qwen3-1.7B}
& 33.3\basex{0.0} & 52.0\basex{0.0} & 24.4\basex{0.0} & 44.2\basex{0.0} & 51.7\basex{0.0} & 41.1\basex{0.0} \\
\hspace*{1em}+ SFT
& 34.1\uprise{0.8} & 36.1\down{15.9} & 13.5\down{10.9} & 42.7\down{1.5} & 39.0\down{12.7} & 33.1\down{8.0} \\
\hspace*{1em}+ OPSD
& 35.5\uprise{2.2} & 45.8\down{6.2} & 23.5\down{0.9} & 46.7\uprise{2.5} & 47.4\down{4.3} & 39.8\down{1.3} \\
\hspace*{1em}+ GRPO
& 37.3\uprise{4.0} & 54.3\uprise{2.3} & 24.1\down{0.3} & 51.6\uprise{7.4} & 59.1\uprise{7.4} & 45.3\uprise{4.2} \\
\hspace*{1em}+ RuscaRL
& 39.5\uprise{6.2} & 54.2\uprise{2.2} & 23.7\down{0.7} & \second{55.2}\uprise{11.0} & \best{61.5}\uprise{9.8} & 46.8\uprise{5.7} \\
\hspace*{1em}+ HeRL
& \best{41.6}\uprise{8.3} & \second{55.8}\uprise{3.8} & \best{27.3}\uprise{2.9} & 52.9\uprise{8.7} & 57.3\uprise{5.6} & \second{47.0}\uprise{5.9} \\
\rowcolor{darkrowgreen} \hspace*{1em}+ \textbf{\ours{}}
& \second{41.2}\uprise{7.9} & \best{57.8}\uprise{5.8} & \second{25.6}\uprise{1.2} & \best{55.6}\uprise{11.4} & \second{61.3}\uprise{9.6} & \best{48.3}\uprise{7.2} \\

\midrule

\textbf{Qwen3-4B}
& 49.7\basex{0.0} & 59.5\basex{0.0} & 38.1\basex{0.0} & 62.2\basex{0.0} & 57.5\basex{0.0} & 53.4\basex{0.0} \\
\hspace*{1em}+ SFT
& 47.0\down{2.7} & 48.1\down{11.4} & 26.5\down{11.6} & 59.1\down{3.1} & 48.3\down{9.2} & 45.8\down{7.6} \\
\hspace*{1em}+ OPSD
& 53.7\uprise{4.0} & 51.4\down{8.1} & 33.1\down{5.0} & 65.2\uprise{3.0} & 52.5\down{5.0} & 51.2\down{2.2} \\
\hspace*{1em}+ GRPO
& 57.8\uprise{8.1} & 62.7\uprise{3.2} & 39.1\uprise{1.0} & 72.4\uprise{10.2} & 67.9\uprise{10.4} & 60.0\uprise{6.6} \\
\hspace*{1em}+ RuscaRL
& \second{62.6}\uprise{12.9} & \second{64.7}\uprise{5.2} & \second{39.3}\uprise{1.2} & \second{75.5}\uprise{13.3} & \second{70.6}\uprise{13.1} & \second{62.5}\uprise{9.1} \\
\hspace*{1em}+ HeRL
& 60.1\uprise{10.4} & 64.6\uprise{5.1} & \second{39.3}\uprise{1.2} & 71.0\uprise{8.8} & 69.2\uprise{11.7} & 60.8\uprise{7.4} \\
\rowcolor{darkrowgreen} \hspace*{1em}+ \textbf{\ours{}}
& \best{63.1}\uprise{13.4} & \best{65.6}\uprise{6.1} & \best{39.7}\uprise{1.6} & \best{76.8}\uprise{14.6} & \best{72.6}\uprise{15.1} & \best{63.6}\uprise{10.2} \\

\bottomrule
\end{tabular}}
\label{tb:qwen3_judge_results}
\end{table*}

\subsection{Results on Other Models}\label{app:other_models}
To assess whether the effectiveness of \ours{} generalizes beyond the base Qwen3-1.7B and Qwen3-4B models, we further evaluate it on an instruction-tuned model (\textbf{Qwen3-4B-Instruct}), a larger model (\textbf{Qwen3-8B}), and a model from a different family (\textbf{Llama3.2-3B-Instruct}) \citep{DBLP:journals/corr/abs-2407-21783}. As shown in Table~\ref{tb:other_models}, \ours{} consistently outperforms both the initial model and GRPO across all model backbones and benchmarks. On RaR-Medicine, \ours{} improves over the initial models by $11.1$, $11.9$, and $22.1$ points, exceeding GRPO by $2.8$, $2.1$, and $10.3$ points on Qwen3-4B-Instruct, Qwen3-8B, and Llama3.2-3B-Instruct, respectively. On HealthBench, \ours{} yields gains of $5.3$, $8.5$, and $10.5$ points over the corresponding initial models and surpasses GRPO by $1.2$, $2.6$, and $3.3$ points. The particularly large improvement on Llama3.2-3B-Instruct demonstrates that the benefits of criterion-level correction transfer beyond the Qwen model family.

\begin{table*}[!t]
\caption{Experimental results ($\%$) on additional model backbones using Qwen3-32B as judge.}
\vspace{0.4em}
\centering
\renewcommand{\arraystretch}{1.25}
\resizebox{0.6\textwidth}{!}{
\begin{tabular}{ll|cc}
\toprule
\textbf{Model} & \textbf{Method} & \textbf{RaR-Medicine} & \textbf{HealthBench} \\
\midrule

\multirow{3}{*}{Qwen3-4B-Instruct}
& Initial & 56.6\basex{0.0} & 62.8\basex{0.0} \\
& GRPO & \second{64.9}\uprise{8.3} & \second{66.9}\uprise{4.1} \\
& \ours{} & \best{67.7}\uprise{11.1} & \best{68.1}\uprise{5.3} \\
\midrule

\multirow{3}{*}{Qwen3-8B}
& Initial & 61.3\basex{0.0} & 62.6\basex{0.0} \\
& GRPO & \second{71.1}\uprise{9.8} & \second{68.5}\uprise{5.9} \\
& \ours{} & \best{73.2}\uprise{11.9} & \best{71.1}\uprise{8.5} \\
\midrule

\multirow{3}{*}{Llama3.2-3B-Instruct}
& Initial & 39.9\basex{0.0} & 44.1\basex{0.0} \\
& GRPO & \second{51.7}\uprise{11.8} & \second{51.3}\uprise{7.2} \\
& \ours{} & \best{62.0}\uprise{22.1} & \best{54.6}\uprise{10.5} \\

\bottomrule
\end{tabular}}
\label{tb:other_models}
\end{table*}

\begin{figure}[t]
    \centering
    \includegraphics[width=1\linewidth]{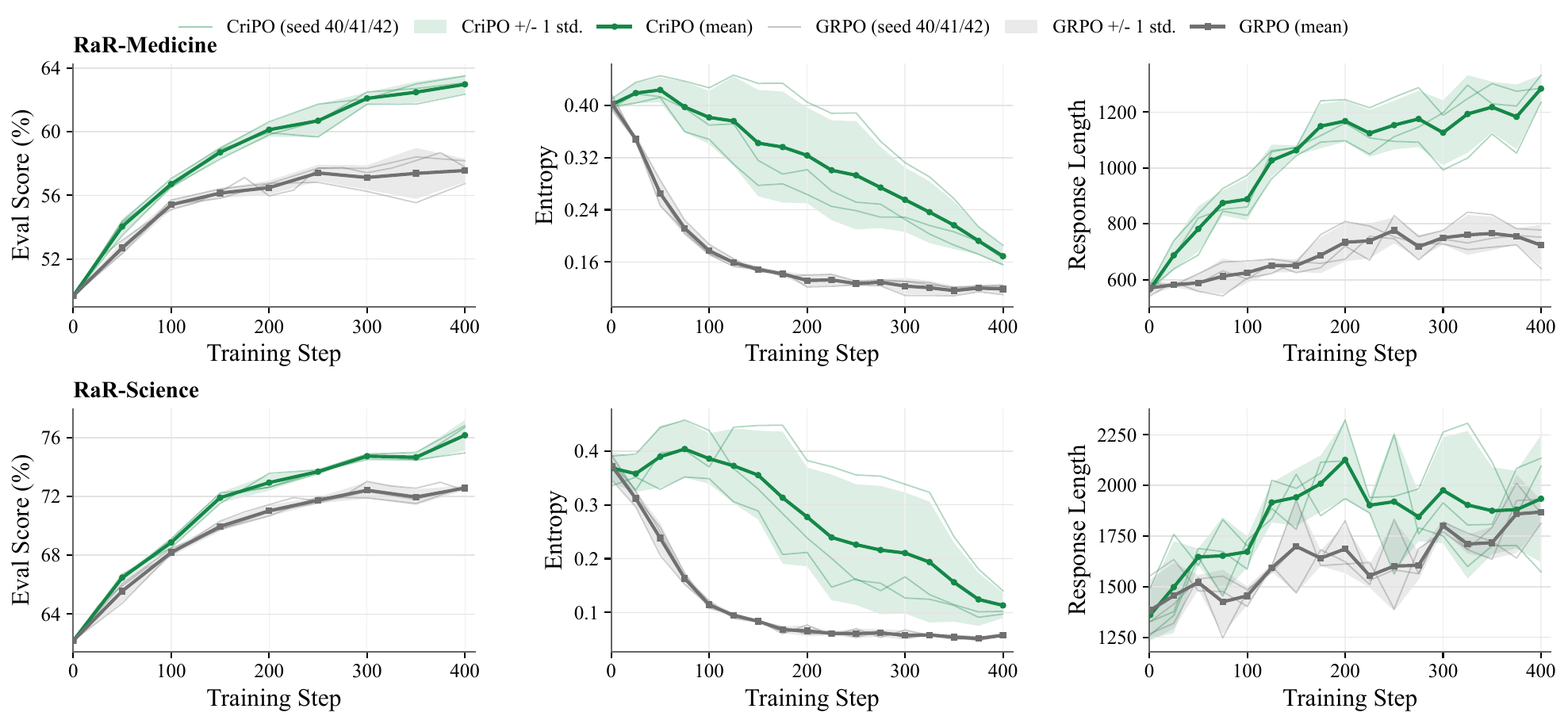}
    \caption{\textbf{Three-seed training dynamics on Qwen3-4B.} We compare \ours{} and GRPO on RaR-Medicine (top) and RaR-Science (bottom) using seeds 40, 41, and 42. Thin lines denote individual runs, thick lines denote the seed mean, and shaded regions show mean $\pm$ one standard deviation. Columns report evaluation score, policy entropy, and response length.}
    \label{fig:seed}
\end{figure}

\subsection{Multi-Seed Results}\label{app:seeds}
Since a complete training run requires substantial computation time (see Appendix~\ref{app:compute_cost}), the main tables report one run for each model--domain configuration. As a sanity check for training variability, we repeat both \ours{} and GRPO with three seeds (40, 41, and 42) using Qwen3-4B on RaR-Medicine and RaR-Science. Figure~\ref{fig:seed} presents the individual trajectories together with their seed mean and $\pm 1$ standard-deviation band.
The evaluation-score trajectories are consistent across seeds: on both domains, every \ours{} run converges to a higher score than the corresponding GRPO runs, and the uncertainty bands are narrow relative to the gap between the two methods in the later stages of training. Moreover, the three-seed means confirm the performance advantage of \ours{} over GRPO observed in the single-seed main results, indicating that the reported gains are not driven by an anomalously favorable run. The other training statistics exhibit the same qualitative behavior observed in the main analysis: \ours{} maintains higher entropy for most of training and generally produces longer responses than GRPO. Overall, these experiments serve as a sanity check that the performance advantage of \ours{} over GRPO is robust to seed variation.

\subsection{Comparison of OPSD Loss Objectives}
\label{app:opsd_variants}

We compare four loss objectives for the standalone OPSD baseline, using the full rubric as the privileged information $\xi$ in Eq.(\ref{eq:opsd}). Let $p_t=\pi_\theta(\cdot\mid x,y_{<t})$ be the student distribution, $q_t=\operatorname{sg}\!\left(\pi_\theta(\cdot\mid x,\xi,y_{<t})\right)$ the self-teacher distribution, and $m_t=(p_t+q_t)/2$. Three objectives use full-distribution, KL-style losses at each token position:
\begin{align}
\ell_t^{\mathrm{KLFKL}} &= D_{\mathrm{KL}}\!\left(q_t\middle\|p_t\right), \label{eq:opsd_klfkl}\\
\ell_t^{\mathrm{KLRKL}} &= D_{\mathrm{KL}}\!\left(p_t\middle\|q_t\right), \label{eq:opsd_klrkl}\\
\ell_t^{\mathrm{KLJSD}} &= \tfrac{1}{2}D_{\mathrm{KL}}\!\left(p_t\middle\|m_t\right)+\tfrac{1}{2}D_{\mathrm{KL}}\!\left(q_t\middle\|m_t\right). \label{eq:opsd_kljsd}
\end{align}
Here KLFKL is forward KL, KLRKL is the reverse KL shown in Eq.(\ref{eq:opsd}), and KLJSD is Jensen--Shannon divergence. We also test policy-gradient-style reverse KL (PGRKL), whose surrogate uses only the log probabilities of the sampled token $y_t$:
\begin{equation}
\ell_t^{\mathrm{PGRKL}}=
\operatorname{sg}\!\left[\log p_t(y_t)-\log q_t(y_t)\right]\log p_t(y_t).
\label{eq:opsd_pgrkl}
\end{equation}
For each variant, the OPSD objective is the expectation of $|y|^{-1}\sum_{t=1}^{|y|}\ell_t$ over on-policy rollouts, as in Eq.(\ref{eq:opsd}). At a fixed prefix, with $y_t\sim p_t$ and the sampled token held fixed during backpropagation, the expected gradient of the PGRKL surrogate equals that of reverse KL. Unlike the three full-distribution losses, PGRKL requires only student and teacher log probabilities for the sampled tokens, without obtaining probabilities over the vocabulary.

Table~\ref{tab:opsd_kl} reports results for Qwen3-4B trained on RaR-Medicine. KLFKL achieves the highest RaR-Medicine score, $53.7$, improving over the initial model by $4.0$ points and exceeding KLRKL, KLJSD, and PGRKL by $1.7$, $2.2$, and $5.4$ points, respectively. On the cross-domain HealthBench evaluation, KLFKL ties with KLRKL for the highest OPSD score of $51.4$, but all four variants fall $8.1$--$8.6$ points below the initial model's $59.5$. This drop suggests that standalone OPSD fits the training-domain rubrics too closely and transfers poorly to rubrics from a different domain. We therefore implement KLFKL for the OPSD baseline in both main-result tables (Tables~\ref{tb:main_results} and~\ref{tb:qwen3_judge_results}). We also ablate the loss type within \ours{}; see Appendix~\ref{app:ablation} for details.

\begin{table*}[ht]
\caption{Comparison of OPSD loss objectives using Qwen3-32B as judge.}
\vspace{0.4em}
\centering
\renewcommand{\arraystretch}{1.25}
\resizebox{0.6\textwidth}{!}{
\begin{tabular}{ll|cc}
\toprule
\textbf{Model} & \textbf{Method} & \textbf{RaR-Medicine} & \textbf{HealthBench} \\
\midrule
\multirow{5}{*}{Qwen3-4B}
& Initial & 49.7\basex{0.0} & 59.5\basex{0.0} \\
& OPSD-KLFKL & \best{53.7}\uprise{4.0} & \best{51.4}\down{8.1} \\
& OPSD-KLRKL & \second{52.0}\uprise{2.3} & \best{51.4}\down{8.1} \\
& OPSD-KLJSD & 51.5\uprise{1.8} & 50.9\down{8.6} \\
& OPSD-PGRKL & 48.3\down{1.4} & \second{51.2}\down{8.3} \\
\bottomrule
\end{tabular}}
\label{tab:opsd_kl}
\end{table*}

\subsection{OOD Generalization on Instruction Following.}\label{app:ood}

To evaluate whether \ours{} preserves general instruction-following ability beyond its rubric-based training domains, we conduct out-of-domain evaluation on IFEval and IFBench. As shown in Table~\ref{tab:ood_if}, both \ours{} variants improve Qwen3-1.7B on the two benchmarks: medicine training yields gains of $0.4$ and $2.0$ points, while science training yields larger gains of $1.3$ and $3.3$ points on IFEval and IFBench, respectively. On Qwen3-4B, the changes are mixed but remain modest: medicine training decreases IFEval and IFBench by $0.6$ and $0.4$ points, whereas science training decreases IFEval by $1.3$ points but improves IFBench by $2.0$ points. Overall, these results suggest that \ours{} largely preserves instruction-following performance on these two benchmarks.

\begin{table*}[ht]
\caption{\textbf{OOD evaluation ($\%$) on instruction-following benchmarks.}
\ours{} (Medicine) and \ours{} (Science) are trained on RaR-Medicine and RaR-Science, respectively. Arrows indicate changes over the corresponding initial model.}
\vspace{0.4em}
\centering
\renewcommand{\arraystretch}{1.25}
\resizebox{0.55\textwidth}{!}{
\begin{tabular}{ll|cc}
\toprule
\textbf{Model} & \textbf{Method} & \textbf{IFEval} & \textbf{IFBench} \\
\midrule

\multirow{3}{*}{Qwen3-1.7B}
& Initial & 69.7 & 18.7 \\
& \ours{} (Medicine) & 70.1\uprise{0.4} & 20.7\uprise{2.0} \\
& \ours{} (Science) & 71.0\uprise{1.3} & 22.0\uprise{3.3} \\

\midrule

\multirow{3}{*}{Qwen3-4B}
& Initial & 80.6 & 25.7 \\
& \ours{} (Medicine) & 80.0\down{0.6} & 25.3\down{0.4} \\
& \ours{} (Science) & 79.3\down{1.3} & 27.7\uprise{2.0} \\

\bottomrule
\end{tabular}}
\label{tab:ood_if}
\end{table*}

\subsection{Adaptation to DAPO}\label{app:dapo}

We further examine whether \ours{} can be incorporated into a stronger RL backbone that explicitly encourages exploration. Specifically, we replace GRPO with DAPO and apply the same criterion-level behavior injection and token-level advantage correction on top of it.
For DAPO, we set the upper importance-ratio clipping bound to $0.28$ and disable KL regularization. To compute the length-penalty score, we set the maximum length to the rollout response-length limit and the cache length to one-quarter of that limit. We retain the token-mean loss aggregation already used in our GRPO implementation.
As shown in Table~\ref{tab:dapo_adaptation}, DAPO outperforms GRPO on RaR-Medicine and the two cross-domain medicine benchmarks, improving RaR-Medicine, HealthBench, and LLMEval-Med by $9.0$, $6.2$, and $5.0$ points, respectively. Adding \ours{} to DAPO yields further gains of $1.6$, $0.3$, and $1.6$ points, showing that \ours{} can also benefit a more exploration-oriented RL backbone.

Because DAPO uses a larger upper clipping bound and disables KL regularization, its stronger policy updates may also increase the risk of drifting from the initial model's general capabilities. We therefore evaluate out-of-domain instruction following on IFEval and IFBench. DAPO decreases performance from $80.6$ to $78.4$ on IFEval and from $25.7$ to $24.3$ on IFBench. In contrast, DAPO+\ours{} retains $79.3$ on IFEval and improves IFBench to $27.3$, outperforming DAPO by $0.9$ and $3.0$ points, respectively. We conjecture that the improved generalization comes from the fine-grained nature of \ours{}: its corrections are concentrated on localized, criterion-relevant tokens instead of broadly altering the entire response distribution. These results highlight the benefit of token-level credit assignment and show that \ours{} can improve both task performance and generalization when adapted to a more exploration-oriented RL objective.

\begin{table*}[t]
\caption{\textbf{Adaptation of \ours{} to DAPO on Qwen3-4B.}
We report performance on medicine benchmarks (judged by Qwen3-32B) and out-of-domain instruction-following (IF) benchmarks. All values are on the percentage scale, and changes are measured against the initial model.}
\vspace{0.4em}
\centering
\renewcommand{\arraystretch}{1.25}
\setlength{\tabcolsep}{4.5pt}
\resizebox{0.8\textwidth}{!}{
\begin{tabular}{l|ccc|cc}
\toprule
\multirow{2}{*}{\textbf{Method}}
& \multicolumn{3}{c|}{\textbf{Medicine}}
& \multicolumn{2}{c}{\textbf{IF}} \\
\cmidrule(lr){2-4} \cmidrule(lr){5-6}
& \textbf{RaR-Medicine}
& \textbf{HealthBench}
& \textbf{LLMEval-Med}
& \textbf{IFEval}
& \textbf{IFBench} \\
\midrule
Initial
& 49.7\basex{0.0} & 59.5\basex{0.0} & 38.1\basex{0.0}
& 80.6\basex{0.0} & 25.7\basex{0.0} \\
GRPO
& 57.8\uprise{8.1} & 62.7\uprise{3.2} & 39.1\uprise{1.0}
& 79.1\down{1.5} & 26.0\uprise{0.3} \\
\ours{}
& 63.1\uprise{13.4} & 65.6\uprise{6.1} & 39.7\uprise{1.6}
& 80.0\down{0.6} & 25.3\down{0.4} \\
DAPO
& 66.8\uprise{17.1} & 68.9\uprise{9.4} & 44.1\uprise{6.0}
& 78.4\down{2.2} & 24.3\down{1.4} \\
DAPO+\ours{}
& 68.4\uprise{18.7} & 69.2\uprise{9.7} & 45.7\uprise{7.6}
& 79.3\down{1.3} & 27.3\uprise{1.6} \\
\bottomrule
\end{tabular}}
\label{tab:dapo_adaptation}
\end{table*}

\subsection{Ablation Studies}\label{app:ablation}

This section examines \ours{} at three levels: the contributions of behavior injection (BI) and advantage correction (AC), design choices within each module, and hyperparameter sensitivity. All results use Qwen3-32B as the evaluation judge. The overall ablation includes Qwen3-1.7B and Qwen3-4B trained on RaR-Medicine and RaR-Science for the corresponding benchmark groups. The sub-module and hyperparameter analyses use Qwen3-4B trained on RaR-Medicine.

\paragraph{Overall Ablation.}
\begin{table*}[t]
\caption{\textbf{Overall ablation of \ours{}.}
BI denotes behavior injection and AC advantage correction; GRPO+BI is \ours{}-U, and GRPO+BI+AC is the full \ours{}.}
\vspace{0.4em}
\centering
\renewcommand{\arraystretch}{1.25}
\resizebox{\textwidth}{!}{
\begin{tabular}{l|ccc|ccc|cc|c}
\toprule
\multirow{2}{*}{\textbf{Model}}
& \multirow{2}{*}{\textbf{GRPO}}
& \multirow{2}{*}{\textbf{BI}}
& \multirow{2}{*}{\textbf{AC}}
& \multicolumn{3}{c|}{\textbf{Medicine}}
& \multicolumn{2}{c|}{\textbf{Science}}
& \multirow{2}{*}{\textbf{Avg.}} \\
\cmidrule(lr){5-7} \cmidrule(lr){8-9}
& & & & \textbf{RaR-Medicine} & \textbf{HealthBench} & \textbf{LLMEval-Med}
& \textbf{RaR-Science} & \textbf{ResearchQA} & \\
\midrule
\multirow{3}{*}{Qwen3-1.7B}
& $\checkmark$ & & & 37.3 & 54.3 & 24.1 & 51.6 & 59.1 & 45.3 \\
& $\checkmark$ & $\checkmark$ & & 39.8 & 55.9 & \textbf{27.0} & 54.6 & \textbf{61.4} & 47.7 \\
& $\checkmark$ & $\checkmark$ & $\checkmark$ & \textbf{41.2} & \textbf{57.8} & 25.6 & \textbf{55.6} & 61.3 & \textbf{48.3} \\
\midrule
\multirow{3}{*}{Qwen3-4B}
& $\checkmark$ & & & 57.8 & 62.7 & 39.1 & 72.4 & 67.9 & 60.0 \\
& $\checkmark$ & $\checkmark$ & & 61.4 & 64.2 & 37.5 & 75.8 & 71.5 & 62.1 \\
& $\checkmark$ & $\checkmark$ & $\checkmark$ & \textbf{63.1} & \textbf{65.6} & \textbf{39.7} & \textbf{76.8} & \textbf{72.6} & \textbf{63.6} \\
\bottomrule
\end{tabular}}
\label{tab:overall_ablation}
\end{table*}

Table~\ref{tab:overall_ablation} compares GRPO, GRPO with behavior injection (\ours{}-U), and the full \ours{}. Adding behavior injection raises the average score from $45.3$ to $47.7$ on Qwen3-1.7B ($+2.4$) and from $60.0$ to $62.1$ on Qwen3-4B ($+2.1$). Adding advantage correction on top of behavior injection further increases the averages to $48.3$ ($+0.6$) and $63.6$ ($+1.5$), respectively. The full method achieves the highest average at both model scales, although individual benchmark scores do not improve uniformly. These results show that both components contribute to the gains over GRPO.

\paragraph{Ablation on Sub-Modules.}
\begin{table*}[t]
\caption{\textbf{Ablation of sub-module designs in \ours{}.}
We report evaluation scores together with the last step's rollout entropy and average response length in thousands of tokens. A dash indicates that the training encounters CUDA OOM due to exploded length.}
\vspace{0.4em}
\centering
\scriptsize
\setlength{\tabcolsep}{4pt}
\renewcommand{\arraystretch}{1.18}
\resizebox{\textwidth}{!}{
\begin{tabular}{lll|cc|cc}
\toprule
\multirow{2}{*}{\textbf{Module}}
& \multirow{2}{*}{\textbf{Sub-Module}}
& \multirow{2}{*}{\textbf{Method}}
& \multicolumn{2}{c|}{\textbf{Eval Performance}}
& \multicolumn{2}{c}{\textbf{Training Dynamics}} \\
\cmidrule(lr){4-5} \cmidrule(lr){6-7}
& & & \textbf{RaR-Medicine} & \textbf{HealthBench}
& \textbf{Entropy} & \textbf{Lengths (k)} \\
\midrule
-- & -- & \ours{} & 63.1 & 65.6 & 0.19 & 1.29 \\
\midrule
\multirow{10}{*}{Behavior Injection}
& \multirow{3}{*}{KL Loss Type}
& KLJSD & 60.1 & 63.1 & 0.11 & 1.26 \\
& & KLRKL & 52.5 & 62.1 & 0.08 & 0.72 \\
& & PGRKL & 64.9 & 69.2 & 0.73 & 2.63 \\
\cmidrule(lr){2-7}
& $\gamma$
& \ours{} w/o selection & 58.3 & 63.5 & 0.11 & 1.02 \\
\cmidrule(lr){2-7}
& \multirow{3}{*}{Dynamic $\beta$}
& \ours{} w/o dynamic & 65.4 & 68.4 & 0.56 & 1.91 \\
& & \ours{}-U & 61.4 & 64.2 & 0.18 & 0.83 \\
& & \ours{}-U w/o dynamic & 60.7 & 64.3 & 0.18 & 0.85 \\
\cmidrule(lr){2-7}
& \multirow{3}{*}{Other Designs}
& Full-rubric teacher & 62.1 & 60.2 & 0.65 & 1.87 \\
& & Rubric-feedback teacher & 62.1 & 65.3 & 0.22 & 1.14 \\
& & Worst-1 rollout & -- & -- & 0.81 & 1.48 \\
\midrule
\multirow{3}{*}{Advantage Correction}
& \multirow{2}{*}{$\tau_{\mathrm{flip}}$}
& All flip & 62.0 & 65.9 & 0.20 & 0.97 \\
& & All mask & 62.6 & 66.4 & 0.21 & 1.10 \\
\cmidrule(lr){2-7}
& Selection Strategy
& Random selection & 61.6 & 65.1 & 0.20 & 0.96 \\
\bottomrule
\end{tabular}}
\label{tab:ablation}
\end{table*}

Table~\ref{tab:ablation} examines the main design choices within behavior injection and advantage correction. For behavior injection, the divergence choice substantially changes both evaluation performance and training dynamics. KLJSD and KLRKL underperform the default forward-KL (KLFKL) design on both benchmarks and exhibit lower entropy and shorter responses. PGRKL obtains higher evaluation scores, but it also sharply increases entropy from $0.19$ to $0.73$ and response length from $1.29$k to $2.63$k, indicating a substantially more aggressive exploration regime. Removing contribution-based token filtering decreases RaR-Medicine and HealthBench by $4.8$ and $2.1$ points, respectively, supporting the need to filter diffuse teacher signals. Removing the dynamic $\beta$ schedule similarly raises the evaluation scores but produces much higher entropy and longer responses; also, \ours{}-U without the schedule suffers from a performance drop on average compared to \ours{}-U, which suggests that dynamic weighting effectively regulates the behavior-injection signal. Among teacher designs, conditioning on the full rubric notably hurts cross-domain generalization on HealthBench and also leads to substantially higher entropy, while the rubric-feedback teacher (i.e., the teacher is conditioned on the criteria that the best rollout does not present) remains closer to \ours{} but is still weaker on RaR-Medicine. These suggest that adding the least necessary criteria information for GRPO is the most stable way. Using the worst rollout instead of the best one suffers from exploded entropy and response length, indicating that distilling on the bad rollout is not reliable.

For advantage correction, applying the same operation to every localized token is less balanced than the selective masking-and-flipping rule: all-flip and all-mask each improve one metric relative to the other, but neither consistently dominates \ours{} across the two benchmarks. More importantly, replacing counterfactual-teacher localization with random token selection (ensuring the same number of selected tokens) lowers RaR-Medicine from $63.1$ to $61.6$ and HealthBench from $65.6$ to $65.1$. This result shows that the benefit does not arise merely from modifying additional token advantages; it depends on locating criterion-relevant tokens.
Moreover, since we apply advantage correction only to negative-advantage rollouts, there exists a theoretical corner case that our method may not cover: a suppressed criterion receives no direct advantage correction when all of its satisfying rollouts have zero advantage. However, we did not observe such a case in the training runs examined, and also, restricting correction to negative-advantage rollouts keeps the counterfactual teacher disjoint from behavior injection on the highest-advantage rollout, preserving at most one additional teacher evaluation per rollout. 
Overall, the ablations show that the sub-module choices jointly control the trade-off among task performance, exploration, and response length.

\paragraph{Hyperparameter Sensitivity.}
\begin{table*}[t]
\caption{\textbf{Hyperparameter sensitivity of \ours{}.}
We report evaluation scores together with the last step's rollout entropy and average response length in thousands of tokens. A dash indicates that the training encounters CUDA OOM due to exploded length.}
\vspace{0.4em}
\centering
\scriptsize
\setlength{\tabcolsep}{4pt}
\renewcommand{\arraystretch}{1.18}
\resizebox{\textwidth}{!}{
\begin{tabular}{lll|cc|cc}
\toprule
\multirow{2}{*}{\textbf{Module}}
& \multirow{2}{*}{\textbf{Sub-Module}}
& \multirow{2}{*}{\textbf{Method}}
& \multicolumn{2}{c|}{\textbf{Eval Performance}}
& \multicolumn{2}{c}{\textbf{Training Dynamics}} \\
\cmidrule(lr){4-5} \cmidrule(lr){6-7}
& & & \textbf{RaR-Medicine} & \textbf{HealthBench}
& \textbf{Entropy} & \textbf{Lengths (k)} \\
\midrule
\multirow{10}{*}{Behavior Injection}
& \multirow{7}{*}{$\gamma$}
& $\gamma=1.0$ (\ours{} w/o selection) & 58.3 & 63.5 & 0.11 & 1.02 \\
& & $\gamma=1.0,\ \beta=0.1$ & 58.0 & 63.3 & 0.10 & 0.99 \\
& & $\gamma=0.95$ (\ours{}) & 63.1 & 65.6 & 0.19 & 1.29 \\
& & $\gamma=0.9$ & -- & -- & 1.65 & 2.01 \\
& & $\gamma=0.9,\ \beta=0.025$ & 62.8 & 64.9 & 0.18 & 1.07 \\
& & $\gamma=0.5$ & -- & -- & 3.11 & 1.90 \\
& & $\gamma=0.5,\ \beta=0.01$ & -- & -- & 1.37 & 1.93 \\
& & $\gamma=0.5,\ \beta=0.005$ & 60.0 & 64.3 & 0.15 & 1.01 \\
\cmidrule(lr){2-7}
& \multirow{3}{*}{$\beta$}
& $\beta=0.1$ & 65.5 & 68.9 & 1.38 & 3.18 \\
& & $\beta=0.05$ (\ours{}) & 63.1 & 65.6 & 0.19 & 1.29 \\
& & $\beta=0.01$ & 58.6 & 63.7 & 0.12 & 1.07 \\
\midrule
\multirow{5}{*}{Advantage Correction}
& \multirow{2}{*}{$\tau_{\mathrm{flip}}$}
& $\tau_{\mathrm{flip}}=1.0$ & 61.8 & 65.3 & 0.21 & 1.28 \\
& & $\tau_{\mathrm{flip}}=0.1$ (\ours{}) & 63.1 & 65.6 & 0.19 & 1.29 \\
\cmidrule(lr){2-7}
& \multirow{3}{*}{Cutoff}
& $\kappa=2048$, $\zeta=10$ (\ours{}) & 63.1 & 65.6 & 0.19 & 1.29 \\
& & $\kappa=5120$ & -- & -- & 0.74 & 1.77 \\
& & $\zeta=5120$ & 62.8 & 65.1 & 0.21 & 1.21 \\
\bottomrule
\end{tabular}}
\label{tab:hyperparameter_sensitivity}
\end{table*}

Table~\ref{tab:hyperparameter_sensitivity} shows that the behavior-injection hyperparameters control a clear trade-off between evaluation performance and training dynamics. Setting $\gamma=1.0$, which removes contribution-based token filtering, lowers performance on both benchmarks. Increasing the strength for \ours{} w/o selection with a larger $\beta$ of 0.1 does not improve its performance, indicating the effectiveness of contribution-based token filtering does not simply stem from increasing the magnitude of the loss.
Reducing $\gamma$ concentrates supervision on fewer high-contribution tokens, but using $\gamma=0.9$ or $0.5$ with a comparatively large distillation coefficient produces substantially higher entropy and longer responses and causes CUDA OOM during training. Reducing the behavior injection loss coefficient $\beta$ stabilizes these settings: $\gamma=0.9,\beta=0.025$ approaches the default performance with similar entropy, whereas the more aggressive $\gamma=0.5,\beta=0.005$ remains weaker. The $\beta$ sweep exhibits the same pattern. A small value of $0.01$ provides insufficient improvement, while $\beta=0.1$ achieves higher evaluation scores at the cost of a large increase in entropy and response length. The default $\gamma=0.95$ and $\beta=0.05$ therefore provide a more balanced operating point.

\begin{figure}[ht]
    \centering
    \includegraphics[width=0.6\linewidth]{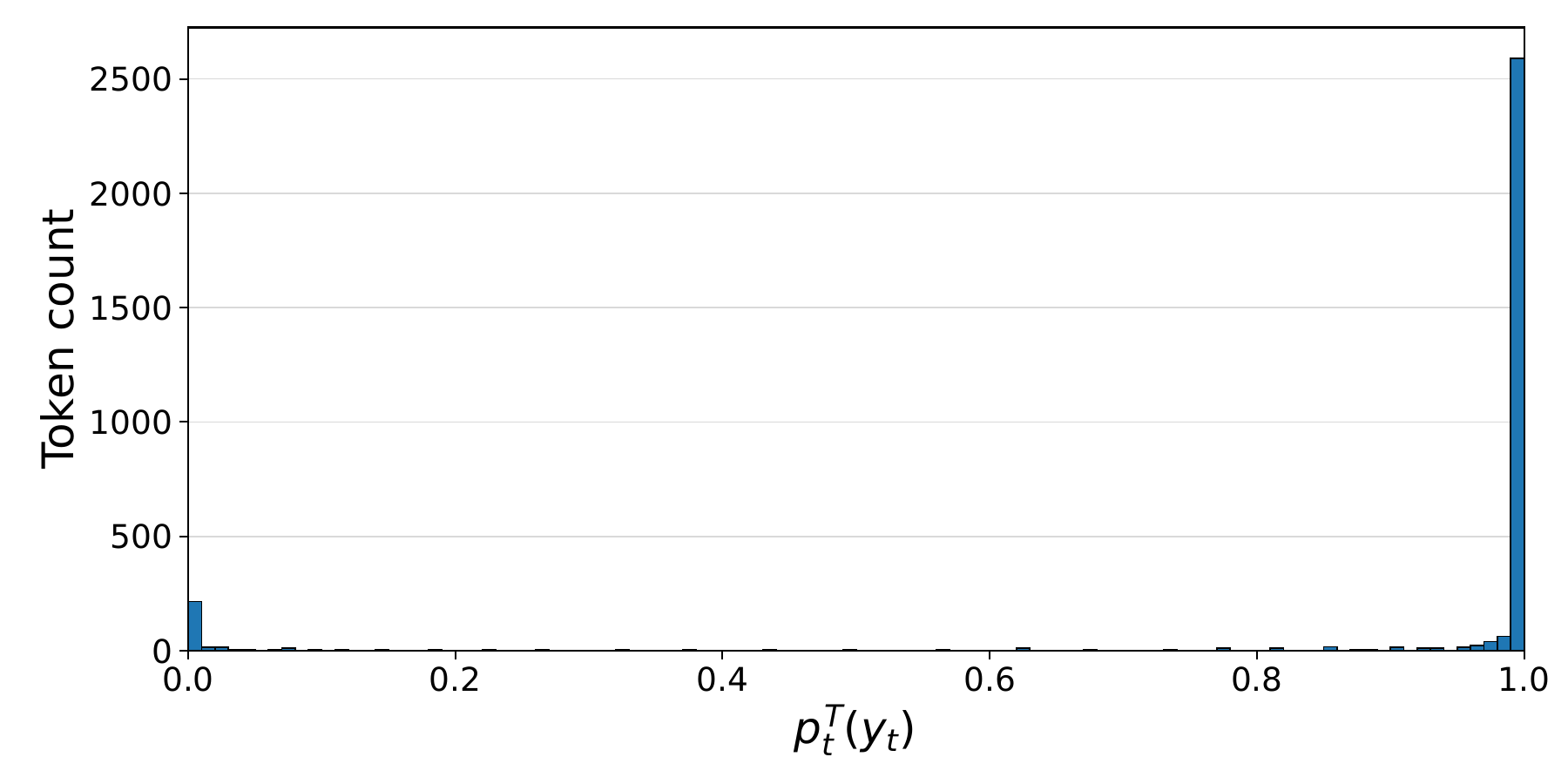}
    \caption{\textbf{Counterfactual-teacher probabilities assigned to sampled tokens.} Histogram of $p^T_{i,t}(y_{i,t})$ during Qwen3-4B training on RaR-Medicine. The 3,344 tokens with probability exactly one are omitted for readability. The concentration near one and smaller low-probability tail motivate the threshold $\alpha=0.1$ for key-token selection.}
    \label{fig:alpha_choice}
\end{figure}

The advantage-correction hyperparameters are less sensitive around the default setting. Increasing $\tau_{\mathrm{flip}}$ from $0.1$ to $1.0$ reduces RaR-Medicine by $1.3$ points and HealthBench by $0.3$ points, supporting a mild positive correction rather than a large flipped advantage. Increasing the selected-token cutoff to $\zeta=5120$ causes a small performance drop, while extending the eligible response length to $\kappa=5120$ substantially raises entropy and response length, resulting in CUDA OOM. Overall, the default $\tau_{\mathrm{flip}}=0.1$, $\kappa=2048$, and $\zeta=10$ is robust and avoids overly aggressive corrections on long responses or large token sets. For the teacher-probability threshold $\alpha$, Figure~\ref{fig:alpha_choice} shows a highly polarized distribution of $p^T_{i,t}(y_{i,t})$ during Qwen3-4B training on RaR-Medicine: most tokens receive probabilities near one, whereas a much smaller subset lies near zero. We therefore set $\alpha=0.1$ to select this low-probability tail, where the counterfactual teacher disfavors the observed token, while excluding the large mass of high-probability tokens.

\section{Case Studies}

\subsection{Case Study on Medical Task}
\label{app:case_study}
\begin{figure}[ht]
    \centering
    \includegraphics[width=1\linewidth]{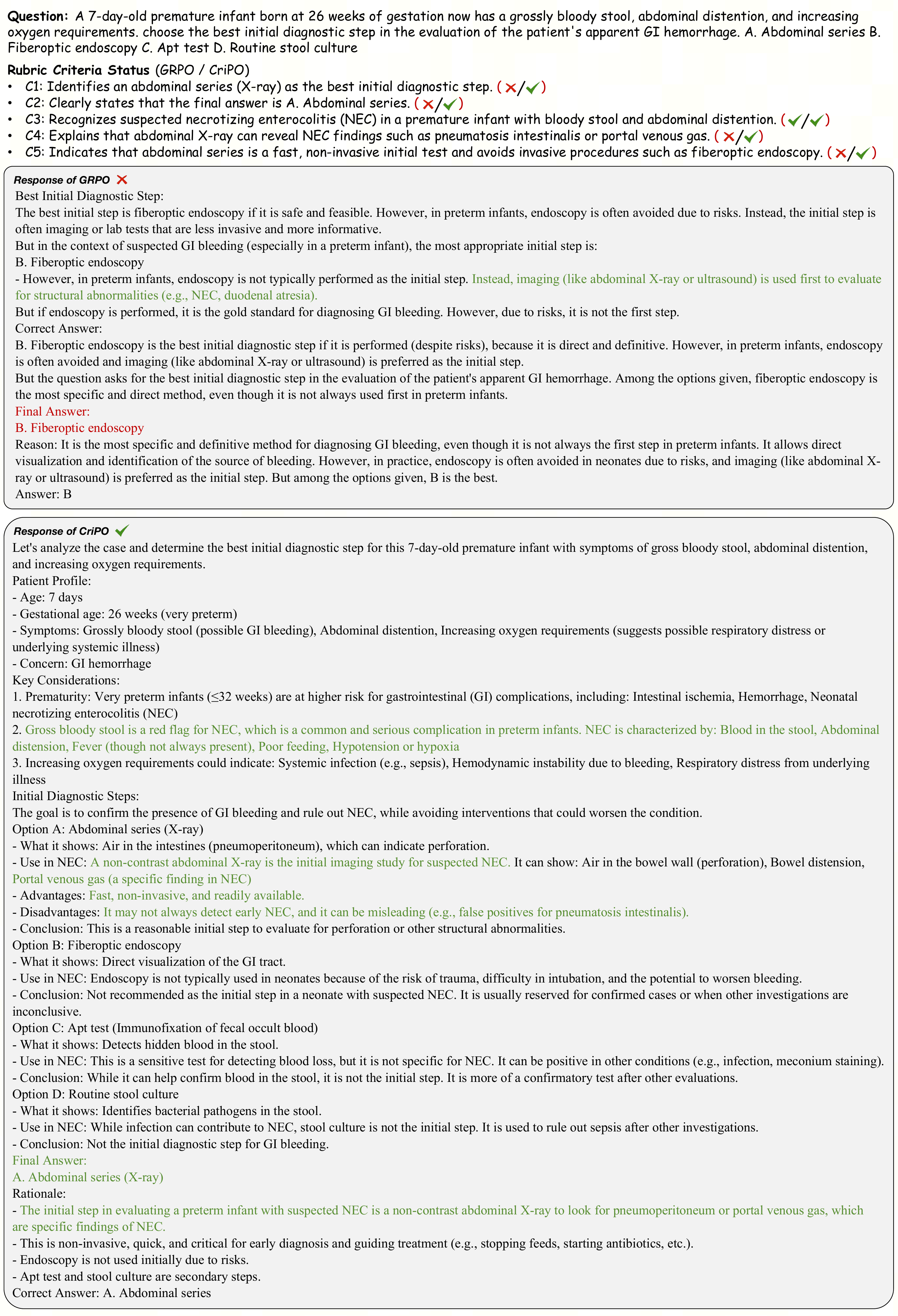}
    \caption{\textbf{Case study on medical reasoning.}
\ours{} produces a more complete reasoning process than GRPO by exploring rubric-relevant clinical cues and diagnostic rationales.}
    \label{fig:case_study}
\end{figure}

Figure~\ref{fig:case_study} compares GRPO and \ours{} on a question about a 7-day-old infant born at 26 weeks of gestation who has grossly bloody stool, abdominal distention, and increasing oxygen requirements. The best initial diagnostic step is option A, an abdominal series, given the concern for necrotizing enterocolitis (NEC). GRPO recognizes NEC as a possibility and even notes that imaging is often preferred before endoscopy in preterm infants; accordingly, it satisfies the rubric criterion for recognizing the clinical presentation. Nevertheless, it repeatedly selects option B, fiberoptic endoscopy, because it treats direct visualization as the more definitive way to evaluate gastrointestinal bleeding. This final choice conflicts with its own caution about the risks of endoscopy and leaves the other four criteria unmet: GRPO neither selects and clearly states the correct answer nor provides NEC-specific radiographic findings or a consistent rationale for starting with a fast, non-invasive test.

By contrast, \ours{} connects the infant's extreme prematurity and clinical signs to suspected NEC, compares the diagnostic options, and selects the abdominal series as the first step. It explains that abdominal radiography can reveal findings such as portal venous gas and bowel abnormalities, while providing a rapid, non-invasive assessment. It also explains why fiberoptic endoscopy is inappropriate as an initial test in this premature neonate because of its procedural risks. The response ends with the explicit answer ``A. Abdominal series,'' satisfying all five criteria shown in the figure. This case illustrates how the additional response length can contain criterion-relevant diagnostic reasoning that supports the final decision.

\subsection{Case Study of Token Selection for Suppressed Criteria.}
\label{app:suppressed_token_cases}

To characterize the suppressed criteria encountered in practice, Table~\ref{tab:criterion_token_distribution} groups them into seven categories and reports a representative criterion and descriptive token-selection statistics for each. The analysis covers 1,641 suppressed-criterion rollout assignments on the testing set of RaR-Medicine using Qwen3-4B. A rollout can be associated with suppressed criteria from multiple categories, so the category rows do not represent disjoint sets of rollouts. These criteria extend beyond medical facts to conciseness, avoidance, reasoning, safety, answer presentation, and style; the table reports their counts and descriptive token-selection statistics. Note that avoidance criteria defined by the absence of particular content pose a challenge for token localization: a response satisfying such a criterion may contain no token that directly expresses the omission; see Bad Case 2 in Table~\ref{tab:suppressed_token_cases}.

\begin{table*}[t]
\caption{\textbf{Statistics for different types of suppressed criteria.} "\#Criteria" reports the number of criteria that occurred, and "\#Tokens" reports the mean number of selected tokens per rollout--criterion-category pair. The five abbreviated columns report the proportions of formatting/structural boundaries (F), knowledge/semantic content (K), logical/discourse connectors (L), grammatical function words (G), and numbers/option labels (N).}
\vspace{0.4em}
\centering
\scriptsize
\setlength{\tabcolsep}{3.5pt}
\renewcommand{\arraystretch}{1.28}
\resizebox{\textwidth}{!}{
\begin{tabular}{lp{0.42\textwidth}rrrrrrr}
\toprule
\textbf{Criterion Category}
& \textbf{Representative Criterion}
& \textbf{\#Criteria}
& \textbf{\#Tokens}
& \textbf{F}
& \textbf{K}
& \textbf{L}
& \textbf{G}
& \textbf{N} \\
\midrule
Medical Facts \& Task Content
& ``Highlights the use of corticosteroids, specifically mentioning intravenous methylprednisolone and oral prednisone.''
& 327 & 4.0 & 45\% & 23\% & 18\% & 13\% & 2\% \\
Conciseness \& Redundancy Control
& ``Remains concise and avoids unnecessary detail in the explanation.''
& 178 & 8.4 & 43\% & 20\% & 16\% & 19\% & 2\% \\
Negative \& Avoidance Criteria
& ``Does not mention distinguishing between fine detail and gross movements, which confuses the parvocellular with the magnocellular pathway.''
& 116 & 3.7 & 44\% & 25\% & 18\% & 12\% & 1\% \\
Explanation \& Reasoning
& ``Explains the correlation between upper abdominal pain and bloody diarrhea with the diagnosis.''
& 55 & 4.3 & 45\% & 20\% & 19\% & 15\% & 2\% \\
Safety \& Uncertainty
& ``Mentions that certain conditions may require heightened caution and monitoring during ECT.''
& 15 & 4.1 & 51\% & 19\% & 18\% & 11\% & 2\% \\
Answer Presentation \& Ordering
& ``Includes a clear statement `The final answer is (D)'.''
& 11 & 5.7 & 58\% & 10\% & 20\% & 10\% & 1\% \\
Style \& Formatting
& ``Uses an empathetic tone in discussing sexually transmitted infections to ensure sensitivity.''
& 6 & 7.4 & 32\% & 47\% & 10\% & 11\% & 0\% \\
\bottomrule
\end{tabular}}
\label{tab:criterion_token_distribution}
\end{table*}

\begin{table*}[t]
\caption{\textbf{Representative good and bad cases of suppressed-criterion token localization.}
Selected key tokens are highlighted in their local response spans. $\varnothing$ indicates that no key token is selected.}
\vspace{0.4em}
\centering
\scriptsize
\setlength{\tabcolsep}{2.5pt}
\renewcommand{\arraystretch}{1.15}
\resizebox{\textwidth}{!}{
\begin{tabular}{p{0.1\textwidth}p{0.3\textwidth}p{0.22\textwidth}p{0.12\textwidth}p{0.26\textwidth}}
\toprule
\textbf{Case Type} & \textbf{User Query} & \textbf{Suppressed Criterion} & \textbf{Key Token(s)} & \textbf{Rollout Span with Key Token(s)} \\
\midrule
\textbf{Good Case 1}
& What drug used in the treatment of osteoporosis acts by both decreasing bone resorption and inducing new bone formation?
& Provides a concise explanation without unnecessary details.
& \texttt{\#\#\#}, \texttt{Explanation}
& RANKL inhibitor---specifically Denosumab. \textcolor{mygreen}{\textbf{\texttt{\#\#\#}}} \textcolor{mygreen}{\textbf{\texttt{Explanation}}}: Denosumab is a monoclonal antibody that targets RANKL. \\
\addlinespace
\textbf{Good Case 2}
& According to the modified Jones criteria, what is one of the minor criteria for the diagnosis of Rheumatic Fever?
& Mentions other minor criteria like fever and certain inflammatory blood test results.
& \texttt{Other}, \texttt{Fever}
& \textcolor{mygreen}{\textbf{\texttt{Other}}} minor criteria include:\newline
-- \textcolor{mygreen}{\textbf{\texttt{Fever}}}\newline
-- Macrocyclic erythema (a type of skin rash)\newline
-- Subcutaneous nodules\newline
-- Choreiform movements\newline
-- Elevated ESR or CRP \\
\addlinespace
\textbf{Good Case 3}
& What is the appropriate treatment for a child with pauci-immune crescentic glomerulonephritis?
& Highlights the use of corticosteroids, specifically mentioning intravenous methylprednisolone and oral prednisone.
& \texttt{Pred}
& \textcolor{mygreen}{\textbf{\texttt{Pred}}}nisone (oral or intravenous) is often used to reduce inflammation. \\
\addlinespace
\textbf{Good Case 4}
& A 35-year-old sexually active male presents with a painless penile vesicle and inguinal lymphadenopathy. The infection has been definitively diagnosed, involving an organism known to exist in both extracellular and intracellular forms. What is the causative organism?
& References the patient's age and sexual activity as context for the infection.
& \texttt{3}
& Clues in the question:
– \textcolor{mygreen}{\textbf{\texttt{3}}}5-year-old sexually active male \\
\midrule
\textbf{Bad Case 1}
& A 32-year-old man has progressive severe headaches, seizures, projectile vomiting, and low-grade fever after otitis media and mastoiditis. What is it?
& Points out that the patient's age can be relevant to the differential diagnosis.
& \texttt{are}
& Otitis media and mastoiditis \textcolor{myred}{\textbf{\texttt{are}}} ear infections that can ascend into the middle cranial fossa and spread to the meninges. \\
\addlinespace
\textbf{Bad Case 2}
& A 23-year-old man experienced unexpected blood loss during a donor nephrectomy and was subsequently transfused with O negative packed red blood cells. He then developed generalized pruritus, expiratory wheezing, and multiple pink, edematous wheals. Given his history of mild recurrent infections as a child, what underlying immunological condition is most likely responsible for these symptoms?
& Does not mention Selective IgA Deficiency as the likely immunological condition.
& $\varnothing$
& The most likely underlying immunological condition is C3 deficiency. \textcolor{myred}{(No key token is selected)} \\
\bottomrule
\end{tabular}}
\label{tab:suppressed_token_cases}
\end{table*}

Table~\ref{tab:suppressed_token_cases} gives concrete examples of what the selected tokens capture. In Good Cases 1--2, structural and discourse markers (\texttt{\#\#\#}, \texttt{Explanation}, and \texttt{Other}) point to an explanatory span or list, while \texttt{Fever} marks relevant medical content. In Good Case 3, the wordpiece \texttt{Pred} in \texttt{Prednisone} identifies more specific content associated with the respective criteria, and in Good Case 4, the selected digit 3 in "35-year-old" marks the patient-age information explicitly referenced by the criterion, illustrating localization of a numerical detail within a relevant response span. We also demonstrate two types of bad cases. The generic word \texttt{are} is selected even though it does not convey the criterion about the patient's age; for the absence-based criterion, no token is selected from a response that omits the specified diagnosis, so that it cannot be repaired by advantage correction. These examples suggest that counterfactual-teacher probabilities can highlight criterion-related cues, but the selected tokens do not always faithfully represent the criterion.

\section{Pseudo-Code of \ours{}}

Algorithm \ref{alg:cripo} shows the pseudo-code of \ours{}.

\begin{algorithm}[ht]
\caption{\ours{}: \oursfull{}}
\label{alg:cripo}
\begin{algorithmic}[1]
\Require Policy $\pi_\theta$; rubric criteria $\mathcal{C}$; rollout number $G$; hyperparameters $\gamma, \alpha, \tau_{\mathrm{flip}}, \beta, \kappa, \zeta$
\Repeat
    \State Sample rollouts $\{y_i\}_{i=1}^{G}$ for prompt $x$, evaluate them with rubric criteria $\mathcal{C}$
    \State Compute rewards $\{R_i\}$, group advantages $\{A_i\}$ and identify Criteria $\mathcal{C}_u,\mathcal{C}_s$
    \State Initialize token advantages $\tilde{A}_{i,t}\leftarrow A_i$ and OPSD loss $\mathcal{L}_{\mathrm{OPSD}}\leftarrow 0$

    \Statex \hspace{0.5em}{\color{blue}$\triangleright$ \textbf{Behavior Injection for Unexplored Criteria}}
    \If{$\mathcal{C}_u\neq\emptyset$}
        \State Select the highest-advantage rollout $y^*$, build teacher and compute forward-KL $d_t$
        \State Select tokens $\mathcal{T}^u$ following Eq.(\ref{eq:kl_selection}) and compute $\mathcal{L}_{\mathrm{OPSD}}$ following Eq.(\ref{eq:behavior_injection})
    \EndIf
    
    \Statex \hspace{0.5em}{\color{blue}$\triangleright$ \textbf{Advantage Correction for Suppressed Criteria}}
    \If{$\mathcal{C}_s\neq\emptyset$}
        \For{each rollout $y_i$ with $A_i < 0$, $|y_i|\leq\kappa$, and $\mathcal{C}_s^i\neq\emptyset$}
            \State Build one counterfactual teacher using all criteria in $\mathcal{C}_s^i$
            \State Select at most $\zeta$ lowest-teacher-probability tokens from Eq.(\ref{eq:flip_select}) as $\mathcal{T}_i^s$
            \State Compute the corrected token-level advantage $\tilde{A}_{i,t}$ following Eq.(\ref{eq:corrected_advantage})
        \EndFor
    \EndIf
    \State Compute $\mathcal{L}_{\mathrm{GRPO}}$ following Eq.(\ref{eq:grpo_flip})

    \State Compute $\mathcal{L}_{\mathrm{\ours{}}}$ following Eq.(\ref{eq:cripo}) and update $\theta$
\Until{converged}
\end{algorithmic}
\end{algorithm}

\section{Limitations}

Despite its gains, \ours{} inherits several constraints of rubric-based RL. Its criterion-level signals depend on the quality and coverage of the rubrics and their judgments: missing, ambiguous, or incorrectly scored criteria can distort the identification of unexplored and suppressed criteria, and consequently misdirect behavior injection or advantage correction. Our experiments use curated rubrics in medicine and science, so the robustness of \ours{} to lower-quality rubric settings remains to be established. Furthermore, the rubrics are fixed throughout training, although the policy's strengths and failure modes change. Criteria that were informative early on may become less useful, while newly emerging weaknesses may remain uncovered. We leave the study of \ours{} under low-quality rubrics and the development of adaptive rubrics that evolve with the policy to provide more targeted supervision to future works.

\section{Prompt Templates}

\begin{figure}[ht]
    \centering
    \includegraphics[width=1\linewidth]{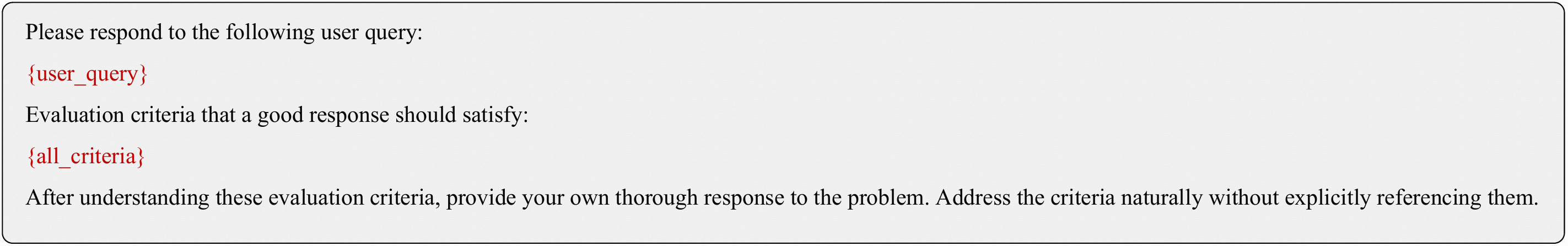}
    \caption{Prompt of the teacher in OPSD baseline and target generation in SFT.}
    \label{fig:full_rubric_teacher_prompt}
\end{figure}

\begin{figure}[ht]
    \centering
    \includegraphics[width=1\linewidth]{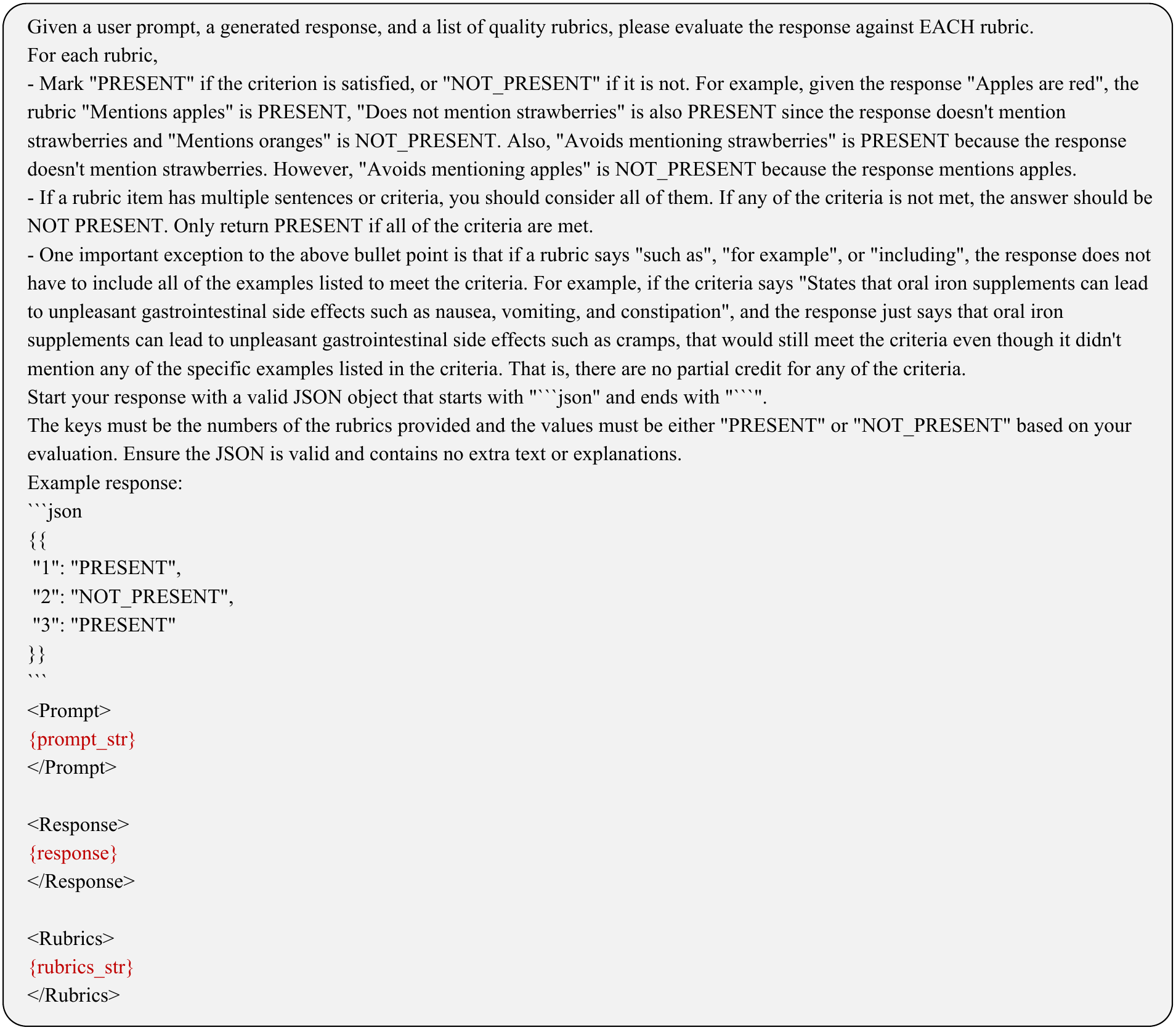}
    \caption{The LLM judge prompt.
The judge evaluates model responses according to a predefined rubric and produces criterion-wise scores for reward computation.}
    \label{fig:judge_prompt}
\end{figure}

The prompt of the teacher in the OPSD baseline, as well as for target generation in SFT, is shown in Figure \ref{fig:full_rubric_teacher_prompt}, and the LLM judge prompt is shown in Figure \ref{fig:judge_prompt}.

\end{document}